\documentclass[11pt]{article}

\usepackage[letterpaper,margin=1.1in]{geometry}
\usepackage[T1]{fontenc}
\usepackage[utf8]{inputenc}
\usepackage{lmodern}
\usepackage{microtype}
\usepackage{amsmath,amssymb}
\usepackage{booktabs}
\usepackage{multirow}
\usepackage{graphicx}
\usepackage{xcolor}
\usepackage{caption}
\usepackage{subcaption}
\usepackage{float}
\usepackage{algorithm}
\usepackage[noend]{algpseudocode}
\usepackage{tikz}
\usetikzlibrary{positioning,fit,backgrounds,calc}
\usepackage{pgfplots}
\pgfplotsset{compat=1.17}
\usepackage{listings}
\usepackage[colorlinks=true, linkcolor=blublue, citecolor=blublue, urlcolor=blublue]{hyperref}

\definecolor{blublue}{RGB}{30,90,160}
\definecolor{blugray}{RGB}{90,90,95}
\definecolor{ptred}{HTML}{EE4C2C}
\definecolor{ptrose}{HTML}{B97376}
\definecolor{bluteal}{HTML}{3A8EAC}
\definecolor{blubg}{RGB}{238,242,248}
\definecolor{codebg}{RGB}{245,245,247}

\newcommand{\code}[1]{\texttt{\small #1}}
\newcommand{\sectref}[1]{\S\ref{#1}}

\title{\vspace{-1.5em}\textbf{BluTrain: A C++/CUDA Framework for AI Systems}\\[6pt]
\large Robust, Lightweight, and Architecture-General , Built from First Principles}

\author{%
\footnotesize
\begin{tabular}{@{}c@{}}
Adhitya Charan \enspace Adwaid Suresh \enspace Anuj Kumar \enspace Aparna A \enspace Dhanakumar K \\[3pt]
Dharun MS \enspace Dinesh G \enspace Goutham Kumar Reddy K \enspace Harshini V M \enspace Jenifa D \enspace Jona Delcy C A \\[3pt]
Kathirvel S \enspace Killi Uma Maheswara Rao \enspace Kiruthik Kanna M \enspace Kurra Vishnu Sai \enspace Madhumithaa G K \\[3pt]
Navin Kumar V \enspace Ram Charan Golla \enspace Revathi T \enspace Rishikkanth R \enspace Sanjay Krishna MV \enspace Surendra Vendra \\[10pt]
\normalsize Blubridge AI \\ \texttt{research@blubridge.com}
\end{tabular}%
}
\date{}

\begin{document}
\maketitle

\begin{abstract}
\noindent
Progress in deep learning is, at scale, more a matter of systems engineering than
of modelling: the behaviour of a model in training (its throughput, its memory
footprint, and the numerical fidelity of the result) is determined less by the
architecture itself than by how that architecture is expressed on the hardware.
To achieve absolute control over this hardware expression while abstracting away 
systems complexity to make modelling seamless and eliminating the need for repetitive orchestration logic,
BluTrain was architected from first principles as a robust, lightweight, and architecture-general 
training framework in standard C++ and the core CUDA programming model. Every layer is implemented natively: a typed tensor module with 
reverse-mode autograd, a linear-algebra library, a caching allocator, a multi-mode 
distributed-execution module, and an MLIR-based deep-learning compiler. 
In formal evaluations training a 124M-parameter GPT-2 baseline in FP32 on an 8-GPU 6000 Ada system, BluTrain outperforms industry-standard baselines in both throughput (sustaining an average of ~407K tokens/s versus PyTorch's ~395K tokens/s) and memory efficiency (achieving up to a 22\% footprint reduction), while strictly preserving numerical fidelity and converging to a marginally lower final validation loss. With every layer explicitly open to native tuning, the performance ceiling is the framework's own to raise. The CUDA kernels underlying these results are available at \url{https://github.com/BlubridgeAI/HyperKernels}.
\end{abstract}

\section{Introduction}
\label{sec:intro}

The capability of a deep-learning model is realized only through the system that
trains it: the tightly coupled integration of the software framework and physical hardware. 
A training run spends almost all of its time in a small set of
operations, such as dense matrix multiplication, normalization, attention, reductions,
the loss, and the optimizer update, each executed billions of times, and the
rate at which the model learns, the memory it occupies, and the numerical fidelity
of the result are decided by how those operations are expressed on the hardware and
coordinated across devices. Building a model is, in this sense, more an act of
systems engineering than of modelling.

BluTrain is a distributed training framework engineered directly from these constraints.
To ensure that mathematical abstractions map precisely to silicon, the framework is structured as
a contiguous, tightly integrated execution pipeline. From the tensor abstraction,
caching allocators, and reverse-mode autograd engine down to the linear-algebra library,
distributed runtime, hardware-specific operator kernels, and an MLIR-based deep-learning compiler, every component is explicitly co-designed and implemented natively with zero dependencies outside of standard C++ and the core CUDA programming model.

Core engineering principles of this architecture: (i) exercise absolute
control over all software layers to enforce optimal hardware expression; (ii) specialize
computations statically for the specific architecture and specific hardware in use;
(iii) maintain strict numerical fidelity to ensure stable convergence at scale;
and (iv) abstract low-level execution to make modelling seamless, eliminating repetitive orchestration logic.

\section{The BluTrain Architecture}
\label{sec:stack}

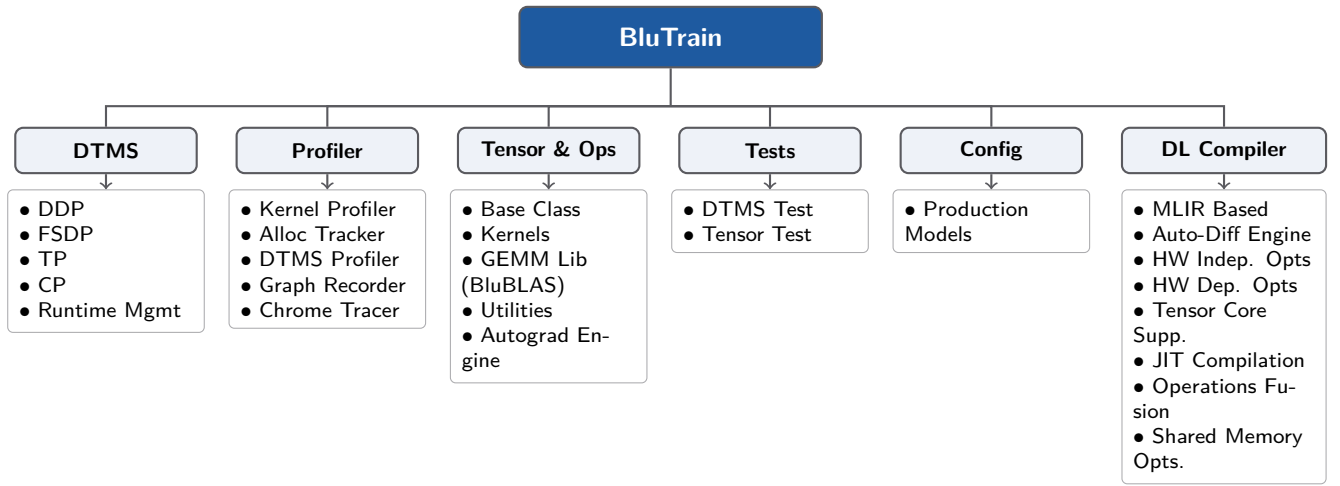
\begin{figure}[hbt!]
\centering
\begin{tikzpicture}[
  font=\sffamily\scriptsize,
  module/.style={draw=blugray, thick, rounded corners=3pt, align=center, fill=blubg, inner sep=1.5mm, minimum width=2.4cm, minimum height=0.6cm},
  root/.style={module, fill=blublue, text=white, font=\sffamily\small\bfseries, minimum width=4cm, minimum height=0.8cm},
  itembox/.style={draw=blugray!50, rounded corners=2pt, align=left, fill=white, font=\sffamily\scriptsize, text width=2.3cm, inner sep=1.5mm, anchor=north},
  line/.style={draw=blugray, thick}
]

\node[module] (tensor) {\textbf{Tensor \& Ops}};
\node[module, left=0.5cm of tensor] (profiler) {\textbf{Profiler}};
\node[module, left=0.5cm of profiler] (dtms) {\textbf{DTMS}};
\node[module, right=0.5cm of tensor] (tests) {\textbf{Tests}};
\node[module, right=0.5cm of tests] (config) {\textbf{Config}};
\node[module, right=0.5cm of config, text width=2.4cm] (nova) {\textbf{DL Compiler}};

\path (dtms.west) -- (nova.east) node[midway, yshift=1.5cm, root] (root) {BluTrain};

\coordinate (split) at ([yshift=-0.5cm]root.south);
\draw[line] (root.south) -- (split);
\draw[line] (split) -| (dtms.north);
\draw[line] (split) -| (profiler.north);
\draw[line] (split) -| (tensor.north);
\draw[line] (split) -| (tests.north);
\draw[line] (split) -| (config.north);
\draw[line] (split) -| (nova.north);

\node[itembox, below=0.2cm of dtms] (dtms_list) {
  $\bullet$ DDP\\
  $\bullet$ FSDP\\
  $\bullet$ TP\\
  $\bullet$ CP\\
  $\bullet$ Runtime Mgmt
};
\draw[line, ->] (dtms.south) -- (dtms_list.north);

\node[itembox, below=0.2cm of profiler] (prof_list) {
  $\bullet$ Kernel Profiler\\
  $\bullet$ Alloc Tracker\\
  $\bullet$ DTMS Profiler\\
  $\bullet$ Graph Recorder\\
  $\bullet$ Chrome Tracer
};
\draw[line, ->] (profiler.south) -- (prof_list.north);

\node[itembox, below=0.2cm of tensor] (tensor_list) {
  $\bullet$ Base Class\\
  $\bullet$ Kernels\\
  $\bullet$ GEMM Lib (BluBLAS)\\
  $\bullet$ Utilities\\
  $\bullet$ Autograd Engine
};
\draw[line, ->] (tensor.south) -- (tensor_list.north);

\node[itembox, below=0.2cm of tests] (tests_list) {
  $\bullet$ DTMS Test\\
  $\bullet$ Tensor Test
};
\draw[line, ->] (tests.south) -- (tests_list.north);

\node[itembox, below=0.2cm of config] (config_list) {
  $\bullet$ Production Models
};
\draw[line, ->] (config.south) -- (config_list.north);

\node[itembox, text width=2.4cm, below=0.2cm of nova] (nova_list) {
  $\bullet$ MLIR Based\\
  $\bullet$ Auto-Diff Engine\\
  $\bullet$ HW Indep. Opts\\
  $\bullet$ HW Dep. Opts\\
  $\bullet$ Tensor Core Supp.\\
  $\bullet$ JIT Compilation\\
  $\bullet$ Operations Fusion\\
  $\bullet$ Shared Memory Opts.
};
\draw[line, ->] (nova.south) -- (nova_list.north);

\end{tikzpicture}\hspace*{0.8cm}
\caption{The modular architecture of the BluTrain framework.}
\label{fig:stack}
\end{figure}

BluTrain is organized as a highly modular ecosystem of natively co-designed components (Figure~\ref{fig:stack}). A centralized configuration drives the instantiation of the model from the foundational tensor module and dictates the distributed topology.

\subsection{Tensor \& Ops Module}
The foundation of the framework is the \textbf{Tensor \& Ops} module, which serves as the primary execution environment. A \code{Tensor} encapsulates its shape, stride, and view offset configurations alongside rich runtime metadata (including datatype, device placement, version counters for in-place mutation tracking, and lazy autograd states), all backed by reference-counted memory storage and tightly coupled to an explicit operator taxonomy. Dense matrix projections are routed through BluBLAS, a native GEMM library (currently hand-tuned for Ampere SM86 and Ada Lovelace SM89 architectures). Gradient computation is handled natively within this module by an integrated reverse-mode Autograd Engine that dynamically constructs a topological computation graph via a dedicated memory arena, driving execution through specialized backward closures.

\subsection{DTMS (Distributed Training Management System)}
Scaling logic is structurally isolated into the DTMS, which orchestrates data, tensor and context parallelisms across the hardware topology. The choice of parallelisms emerges from the need to increase the throughput or train bigger models or enable long context training or a mix of these and is constrained by the interconnect topology. To effectively hide communication latency, the module implements communication-computation overlap by executing collective communications on a dedicated CUDA stream concurrently. This distribution is enabled by an orchestrator that manages the parallel execution of the training run and provides resilience against the faults that can be encountered during the runtime.

\subsection{Profiler and Tests}
To maintain absolute visibility into hardware utilization, execution telemetry is captured via a zero-overhead \textbf{Profiler} module. It bypasses high-level abstractions to provide exact kernel-level timings, nanosecond-resolution memory allocation tracking, and native integration with the NVIDIA Nsight profiling suite (Nsight Systems and Nsight Compute). Furthermore, a dedicated precision logger captures bit-exact cross-framework diffs (including L2 and Infinity norms) against industry-standard frameworks by aggressively dumping full forward I/O and backward gradients to disk. Validation is strictly enforced through an internal \textbf{Tests} module spanning multiple highly compartmentalized test suites, confirming that numeric fidelity is mathematically preserved across all architectural iterations.

\subsection{Deep-Learning Compiler}
To transcend the boundaries of isolated hand-written kernels and optimize the entire computational graph as a unified algebraic structure, the framework integrates a custom MLIR-based~\cite{mlir} deep-learning compiler acting as a just-in-time (JIT) backend. Operating entirely agnostic to high-level neural network semantics, the compiler directly ingests traced forward and backward autograd graphs as pure mathematical operations into a native MLIR dialect. This allows the system to apply compounding layers of global algebraic optimizations before progressively lowering the representation through Linalg, SCF, and Vector abstractions down to highly tuned NVPTX kernels. The compiler explicitly classifies tensor contractions by arithmetic intensity. This classification determines the optimal tensor-core (MMA) intrinsic and two-dimensional warp-grid schedule, driving aggressive shared-memory promotion, software pipelining, and multi-buffering. During execution, an LLVM ORC LLJIT engine and custom ABI adapter bind the compiled kernels directly back to live tensor storage using zero-copy memref descriptors, completely eliminating memory marshaling overhead.

\section{Kernel Design and Optimization}
\label{sec:kernels}

Across the full operator and kernel surface required for a training step (enumerated
in \sectref{sec:stack}), each kernel is implemented natively in highly optimized C++ and CUDA, 
specialized strictly to the shapes, datatypes, and hardware architecture executing it. The advantage
comes not from any individual kernel but from a rigorous optimization discipline applied uniformly 
across both host and device boundaries, governed by a core set of principles:

\begin{itemize}\itemsep2pt
  \item \emph{Algorithmic Co-design:} Hardware-agnostic algorithmic optimizations (e.g., mathematical reformulations) and hardware-dependent optimizations (e.g., vectorization, PTX intrinsics, and shared-memory pipelining) are mutually dependent. An elegant algorithm mapped poorly to the hardware will inevitably be bottlenecked by microarchitectural constraints, while aggressive hardware-dependent optimizations applied to an inefficient algorithm merely execute redundant work faster. Maximum throughput across the entire kernel surface is achieved strictly when high-level algorithmic efficiency is perfectly coupled with low-level microarchitectural hardware optimizations.
  \item \emph{Compile-Time Specialization:} Execution parameters such as datatypes, memory layouts, algorithmic blocking, and hardware microarchitectures are defined as template parameters resolved strictly at compile time. This ensures each kernel collapses into a branch-free, fully unrolled executable tailored for the exact problem geometry, entirely eliminating runtime dispatch overhead and dead branches. A unified C++ template dispatch architecture compiles directly into highly specialized native machine code. This architecture spans the host-device divide, lowering directly into native machine code for the host and specialized execution binaries for the device. This ensures uncompromising hardware expression across heterogeneous environments, seamlessly bridging diverse CPU targets with advanced GPU microarchitectures.
  \item \emph{Roofline Saturation:} Every kernel is fundamentally classified by its arithmetic intensity as either compute-bound or memory-bound, and systematically engineered to saturate its corresponding architectural ceiling. Compute-bound operations (e.g., Attention, GEMM) are optimized to maximize Tensor-Core throughput via aggressive SRAM data reuse and asynchronous DRAM-to-SRAM pipelines (\code{cp.async}), effectively hiding global memory latency behind dense matrix arithmetic. Memory-bound operations (e.g., Normalization, Activations, Reductions, Optimizers) are driven to the device's absolute DRAM speed-of-light via single-pass mathematical formulations, maximum-width vectorized memory transactions (e.g., 128-bit \code{float4} loads), and direct warp-level reductions that entirely bypass shared-memory traffic.
  \item \emph{Uncompromising Numerical Fidelity:} Accumulation order, rounding modes, and precision boundaries are enforced as explicit architectural constraints. The framework strictly utilizes mathematically stable algorithms across the entire kernel surface, entirely avoiding lossy approximations. To enforce this standard, every kernel undergoes rigorous precision validation against industry-standard baselines. This uncompromising discipline ensures that mathematical correctness is never traded for execution speed, enabling the framework to precisely reproduce reference training curves (\sectref{sec:eval}) rather than loosely approximate them.
  \item \emph{Maximum Silicon Utilization:} Sustaining high hardware utilization requires aggressively minimizing host-side latency and eliminating artificial software barriers. By systematically stripping away high-level interpreter overheads, dynamic dispatch bottlenecks, and redundant memory allocations, the framework is engineered to keep the GPU's Streaming Multiprocessors (SMs) as actively engaged as physically possible. The ultimate objective is to translate raw computational capability directly into observable, end-to-end throughput, ensuring the hardware spends its cycles strictly on mathematical execution rather than stalling on host-side coordination.
\end{itemize}

Applied uniformly, this architectural discipline translates directly to high-throughput hardware execution. Profiling telemetry confirms that the execution engine systematically approaches practical hardware limits, achieving near-optimal resource utilization across both compute-intensive and memory-bound regimes. For a comprehensive, shape-specific breakdown of performance metrics (including precise latency comparisons, bandwidth utilization, and detailed execution profiles across the primary computational kernels), refer to Appendix~\ref{app:benchmarks}.

\section{Distributed Execution}
\label{sec:parallel}

While the DTMS abstraction isolates the parallelism strategy from the model definition, realizing high-throughput distributed training requires mapping these theoretical strategies directly to the physical interconnect. This section details the execution of the multi-dimensional parallelization topology across data, tensor, and context execution. By treating distributed execution as a fundamental architectural primitive and strictly controlling the intersection of compute kernels with asynchronous collective communications, the architecture ensures near-linear scaling without exposing synchronization overhead to the critical execution path.

\subsection{Data-Parallel Orchestration}
The data-parallel topology overlaps communication with backward computation through an asynchronous, pipelined data-flow model~\cite{pytorch_ddp} coordinated by lightweight GPU-side stream dependencies rather than host synchronization. Algorithm~\ref{alg:ddp} formalizes this step orchestration. Execution is aggressively decoupled into local accumulation micro-steps and a single synchronizing micro-step. During gradient accumulation, forward and backward propagation execute entirely unhindered, with no distributed communication launched. Synchronization is deferred strictly to the final micro-step. Here, the autograd engine's execution graph dictates the exact emission sequence of parameter gradients; as each gradient is emitted, it is copied into its predefined 25\,MB contiguous memory bucket, and completed buckets are all-reduced asynchronously, overlapping communication with the remaining backward pass.

When a bucket saturates, an asynchronous \code{AllReduce} collective is dispatched on a secondary, high-priority CUDA stream. This mechanism interleaves dense network transfers with the remaining backward matrix multiplications. To preserve compute/communication overlap without stalling the host, the framework inserts GPU-side cross-stream dependencies using low-overhead CUDA events (\code{cudaEventRecord} + \code{cudaStreamWaitEvent}): the communication stream waits on the compute stream before each bucket's all-reduce, and the compute stream waits on the communication stream before the optimizer step. All synchronization is strictly stream-to-stream; the host thread never blocks on a barrier. Extended distributed scaling benchmarks, detailing the impact of varying bucket sizes on \code{AllReduce} latency and overall throughput, are provided in Appendix~\ref{app:ddp}.

\begin{algorithm}[H]
\caption{Distributed Data Parallel Training Step}
\label{alg:ddp}
\small
\begin{algorithmic}[1]
\Require params $P = [p_0, \dots, p_{N-1}]$, world size $W$, bucket cap $C$
\State $grad\_accum\_steps \gets \text{global\_batch} / (B \cdot T \cdot W)$
\State Group $P$ into buckets in gradient-ready order \Comment{bucket $0$ = first ready}
\State Broadcast $P$ from rank 0 \Comment{identical start weights}
\For{$i \gets 0$ \textbf{to} $N - 1$}
  \State Register a hook on $p_i$ \Comment{fired when $\nabla p_i$ is ready}
\EndFor

\For{$micro\_step \gets 0$ \textbf{to} $grad\_accum\_steps - 1$}
  \State $grad\_sync \gets (micro\_step = grad\_accum\_steps - 1)$ \Comment{reduce only on last micro-step}
  \State $loss \gets \text{cross-entropy of } \Call{FORWARD}{x} \text{ against target}$
  \State Reset $bucket.pending$ for all buckets; $next\_bucket \gets 0$
  \State Run backward pass \Comment{fires hook below per gradient}
\EndFor
\State Optimizer step; zero gradients

\Function{HOOK}{$i$} \Comment{on $\nabla p_i$ ready}
  \If{$\neg grad\_sync$}
    \State \textbf{return}
  \EndIf
  \State Write $\nabla p_i$ into its bucket $b$
  \State $bucket[b].pending \gets bucket[b].pending - 1$
  \While{$buckets[next\_bucket].pending = 0$} \Comment{in-order launch gate}
    \State Launch async all-reduce (average) on $buckets[next\_bucket]$
    \State $next\_bucket \gets next\_bucket + 1$
  \EndWhile
\EndFunction

\Function{FINALIZE}{} \Comment{after last gradient}
  \State Wait on all bucket all-reduces
  \State Scatter averaged gradients back to $\nabla p_i$
\EndFunction
\end{algorithmic}
\end{algorithm}

\subsection{Tensor Parallelism}
To push past the per-device memory limits of a single GPU, the architecture natively implements \textbf{Tensor Parallelism}. Dense matrix operations, such as the query-key-value (QKV) projections, attention-output matrices, MLPs, and vocabulary-scale embeddings, are systematically sharded across a defined multi-device execution mesh~\cite{pytorch}. This sharding strictly follows symmetric column- and row-parallel schemes: column-parallel layers produce independent output shards, while row-parallel layers emit partial sums that are immediately synchronized via an \code{AllReduce} collective~\cite{pytorch,megatron}. In the backward pass, gradient reductions are mirrored symmetrically, ensuring that parameter updates remain strictly localized to their respective physical devices while activation gradients propagate upstream.

To prevent collective communications from stalling the compute stream during tensor synchronization, the runtime implements a custom dual-stream \code{AllReduce} overlap protocol, adapting recent asynchronous tensor parallelism paradigms~\cite{async_tp}. Computations are dynamically subdivided into discrete chunks along the token sequence dimension. Execution is asynchronously partitioned across two parallel CUDA queues: the foundational GEMM contractions are dispatched to a primary compute stream, while the resulting collective synchronizations are simultaneously offloaded to a non-blocking NCCL stream. Through precise CUDA event signaling, the \code{AllReduce} transfer for chunk $i$ is hidden entirely behind the matrix multiplication for chunk $i+1$. By explicitly scheduling non-blocking communication kernels concurrently with dense matrix operations, the runtime effectively removing collective latency from the critical path, driving near-linear scaling without sacrificing computational density. Extended tensor-parallel scaling benchmarks, detailing empirical throughput comparisons against Megatron-LM and dual-stream \code{AllReduce} overlap profiles, are provided in Appendix~\ref{app:tp}.

\subsection{Context Parallelism}
To scale attention computations across sequences that exceed a single device's memory capacity, the framework implements a native \textbf{Context Parallelism} (CP) mechanism, adapting recent distributed sequence paradigms~\cite{pytorch_cp}. Context Parallelism is architecturally distinct from NVIDIA's tensor-parallel-tied Sequence Parallelism, which partitions activations only across non-tensor-parallel regions (LayerNorm, Dropout) and requires TP as a prerequisite. In contrast, CP operates as a standalone parallelism dimension: it distributes the sequence length itself across the multi-device execution mesh, independent of how the embedding dimensions are sharded. By treating the global sequence as a distributed ring of local chunks, the framework executes the core attention mechanism~\cite{flashattn} iteratively across devices.

To sustain this distributed operation, the architecture passes Key (K) and Value (V) shards across the GPU mesh using a Ring Rotator pipeline, mathematically analogous to Ring Attention paradigms~\cite{ringattn}. The forward pass utilizes a single ring rotator, while the backward pass instantiates two isolated rotators: one to recompute the attention scores and a second to transfer the corresponding gradient tensors. This communication is supported by interchangeable backend collectives, including All-to-All, AllGather, and Peer-to-Peer (P2P) transfers, explicitly configured based on the underlying hardware topology.

To maintain global mathematical equivalence during the distributed forward pass, local attention outputs are incrementally aggregated across the sequence dimension using an iterative block-wise LogSumExp (LSE) merging strategy. Specifically, as KV shards circulate through the GPU mesh, the partial attention outputs (\code{block\_out}) and corresponding LSE values (\code{block\_lse}) from each local iteration are combined with the running global state (\code{out}, \code{lse}) via a numerically stable sigmoid-scaled update mechanism. Furthermore, to eliminate idle device cycles induced by causal masking, the runtime employs a deterministic load-balancing scheme~\cite{cp_loadbalance} that statically rearranges token chunks immediately at the data-loading boundary, ensuring computational parity across all GPUs globally throughout the network layers. Extended context-parallel scaling benchmarks, detailing throughput and convergence comparisons against PyTorch across the All-to-All, P2P, and AllGather rotators on the RTX 5070 and RTX 6000 Ada testbeds, are provided in Appendix~\ref{app:cp}.

\subsection{Distributed Orchestration}
Sustaining a training run across days or weeks on real hardware demands an operational layer apart from the parallelism strategies above: one that spawns processes and supervises process gangs, detects failures, and recovers from them without interrupting training. At the scale and duration of modern training runs, hardware faults are not exceptional but statistically inevitable, and the established discipline for tolerating them is to checkpoint the full execution state and resume from the last uncorrupted point~\cite{megatron_v2,checknrun}.

A dedicated distributed orchestration substrate provides this through multiple integrated components. An orchestrator daemon assigns GPUs based on NVML-discovered NUMA topology and live device health, excluding any GPU flagged as degraded or unavailable, and seeds each gang's environment with a pre-computed NCCL unique identifier and a private per-job Store, collapsing the multi-rank rendezvous to a single deterministic phase. A launcher forks the process gang with precise NUMA pinning, drains each rank's stdout and stderr into Promtail-watched log files, and broadcasts a NCCL abort signal through the per-job store on teardown, guaranteeing clean collective shutdown on every exit path. A sentinel aggregates telemetry from NVML, DCGM, node\_exporter, and a BMC Redfish poller, capturing GPU temperature, ECC single- and double-bit error deltas, XID event codes, throttle reason bitmasks, NVLink and PCIe bandwidth, CPU RAPL power, NIC error rates, PSU output etc., while tracking a per-rank heartbeat clock over the per-job store.

An analytical layer runs trend analysis on temperature, ECC counters etc., to find patterns surfacing predictive faults before the hardware crosses into hard failure, an approach consistent with large-scale empirical studies of GPU failure signatures~\cite{checknrun,gpufield}. A fault engine classifies every incident against a taxonomy spanning GPU hardware (XID events, NVLink fabric faults, NCCL collective hangs and timeouts, GPU and host OOM, checkpoint corruption and truncation, framework and driver crashes, OS-level process kills, and per-rank straggler divergence) and drives a persisted recovery state machine through gang kill, VRAM drain, UVM reload or full GPU reset, and job requeue from the last checkpoint.

The full recovery history is committed to structured logs, Prometheus metrics, and OTLP traces under a stable trace identifier shared across every attempt. The result is a training infrastructure that, in line with the resilience practices of large-scale training systems~\cite{megatron_v2}, renders the dominant failure modes of long-running multi-GPU training operationally invisible: a run that encounters an ECC double-bit error, a hung collective, a thermal excursion, an OOM kill, or a partial gang exit emerges from recovery at its last uncorrupted checkpoint, the training loop completing despite the operational issues encountered. Checkpoint save and load latency benchmarks for this subsystem, including the asynchronous staging path, are provided in Appendix~\ref{app:ckpt}.

\section{Implementation}
\label{sec:impl}

The architectural constraints of the framework are sustained by a vertically integrated, native runtime environment. The system explicitly eschews high-level framework dependencies in favor of purpose-built, custom infrastructure components designed to maximize continuous hardware utilization.

\paragraph{Native GEMM Library:}
Dense matrix multiplication is served exclusively by \textbf{BluBLAS}~\cite{blublas}, a custom, fully specialized GEMM library written in explicit PTX and CUDA C++. It acts as the foundational compute engine for all dense tensor contractions within the framework, specifically the query-key-value (QKV), attention-output, and MLP projections across all forward and backward execution layouts. Operating directly at the instruction level, the library implements advanced memory-system co-design to ensure continuous L2 cache residency and maximum Streaming Multiprocessor (SM) occupancy, even on highly irregular, low-parallelism gradient shapes. A detailed analysis of its microarchitectural optimizations and complete performance benchmarks are presented in its dedicated technical report.

\paragraph{Horizontal Tensor Fusion:}
To eliminate the host-side overhead typically associated with massive parameter updates, the runtime implements a global multi-tensor horizontal fusion strategy. A single optimizer execution explicitly collapses hundreds of discrete tensor updates into a unified kernel launch. Through an advanced chunk-based load-balancing scheduler, workloads are distributed evenly across the hardware regardless of irregular tensor boundaries. This strategy ensures that all streaming multiprocessors remain saturated without tail-effect stalling, eliminating launch overhead and driving memory-bound operations to nearly 90\% of the physical DRAM bandwidth limit.

\paragraph{Custom Caching Allocator:}
Device memory is managed by a deterministic, self-tuning block-pool allocator engineered for the cyclic memory lifecycles of deep learning. To eliminate internal fragmentation, the allocator bypasses rigid size classes. Instead, it profiles tensor allocation frequencies and historical wasted bytes during the initial forward-backward pass, deriving an optimal alignment configuration dynamically tailored to the active computational graph. Additionally, the runtime executes a targeted cache flush following the 0th step to clear anomalous initialization overhead. This isolates temporary startup buffers from the steady-state memory pool, explicitly releasing the reclaimed device memory back to the CUDA driver.

\paragraph{Data Loader:}
The ingestion pipeline is architected to overlap data movement with computation rather than serializing them (Algorithm~\ref{alg:dataloader}). Token shards are memory-mapped and indexed by a rank-strided cursor, enabling distributed workers to partition the corpus without inter-process coordination. To keep host-to-device (H2D) transfer off the critical path, the loader stages each batch through page-locked (pinned) host buffers and issues an asynchronous \code{cudaMemcpyAsync} on a dedicated copy stream, so the next batch is prefetched into device memory while the current batch is being computed. Cross-stream ordering is enforced with CUDA events: one signals when a batch's H2D copy has landed before the compute stream reads it, and a second prevents the copy stream from overwriting a buffer the consumer is still using. As a result, input transfer is hidden whenever per-step compute exceeds per-step transfer; if pinned allocation is unavailable, the loader falls back to a synchronous path with identical results.

\begin{algorithm}[H]
\caption{Double-Buffered Batch Production}
\label{alg:dataloader}
\small
\begin{algorithmic}[1]
\Require Sorted shards $S = [s_0, \dots, s_{N-1}]$, mini-batch $(B, T)$, rank $r$, world size $W$. Let $BT = B \cdot T$.
\State \textbf{Init:} open $s_0$; $pos \gets BT \cdot r$; $primed \gets \textbf{false}$; $buf \gets 0$
\State Allocate pinned buffers $P_x[0..1], P_y[0..1]$ and device tensors $X[0..1], Y[0..1]$
\State Create copy stream and events

\Function{NEXTBATCH}{}
  \If{pinned allocation failed}
    \State \textbf{return} \textproc{SYNCFALLBACK}
  \EndIf
  \If{$\neg primed$}
    \State \Call{ISSUEPREFETCH}{0}; \Call{ISSUEPREFETCH}{1} \Comment{prime both channels}
    \State $primed \gets \textbf{true}$; $buf \gets 0$
  \Else
    \State Record \texttt{consumer\_done} on default stream
    \State Copy stream waits on \texttt{consumer\_done} \Comment{buffer safe to overwrite}
    \State \Call{ISSUEPREFETCH}{$1 - buf$} \Comment{refill channel returned last call}
  \EndIf
  \State Default stream waits on \texttt{h2d\_done}[$buf$] \Comment{data has landed}
  \State $b \gets (X[buf], Y[buf])$; $buf \gets (buf + 1) \bmod 2$
  \State \textbf{return} $b$
\EndFunction

\Function{ISSUEPREFETCH}{$i$}
  \If{$pos + BT + 1 > |s_{cur}|$}
    \State \textproc{ADVANCESHARD}
  \EndIf
  \State $t \gets \text{data\_ptr} + pos$ \Comment{zero-copy pointer into mmap}
  \State $\text{memcpy}(P_x[i], t, BT)$; $\text{memcpy}(P_y[i], t+1, BT)$ \Comment{host: mmap $\to$ pinned}
  \State $\text{memcpyAsync}(X[i] \gets P_x[i])$; $\text{memcpyAsync}(Y[i] \gets P_y[i])$ on copy stream
  \State Record \texttt{h2d\_done}[$i$] on copy stream
  \State $pos \gets pos + BT \cdot W$ \Comment{global rank stride}
  \If{$pos + BT \cdot W + 1 > |s_{cur}|$}
    \State \textproc{ADVANCESHARD}
  \EndIf
\EndFunction

\Function{ADVANCESHARD}{}
  \State $cur \gets (cur + 1) \bmod N$; open $s_{cur}$; $pos \gets BT \cdot r$
\EndFunction
\end{algorithmic}
\end{algorithm}

\paragraph{Build and execution:}
The entire architecture compiles into a unified, standalone binary without reliance on extensive external toolchains. Training executes strictly as a native C++ process, entirely eliminating the inherent global interpreter locks and runtime overheads associated with Python environments. High-level scripting languages are deliberately relegated strictly to offline analysis, ensuring the training loop remains a pure, unhindered execution engine.

\section{Evaluation}
\label{sec:eval}

To evaluate the correctness and performance characteristics of BluTrain, we model a GPT-2 style decoder-only Transformer~\cite{gpt2} containing 124 million parameters and compare its behavior against an equivalent implementation in PyTorch~\cite{pytorch}.

Our evaluation focuses on the three primary metrics that matter most for a training framework: \emph{numerical fidelity} (convergence quality), \emph{throughput} (tokens/s), and \emph{memory} (footprint). The microbenchmarks in \sectref{sec:kernels} explain the end-to-end speedups that occur.

\subsubsection*{GPT-2 Architecture}
\begin{figure}[H]
    \centering
    \includegraphics[width=0.6\textwidth]{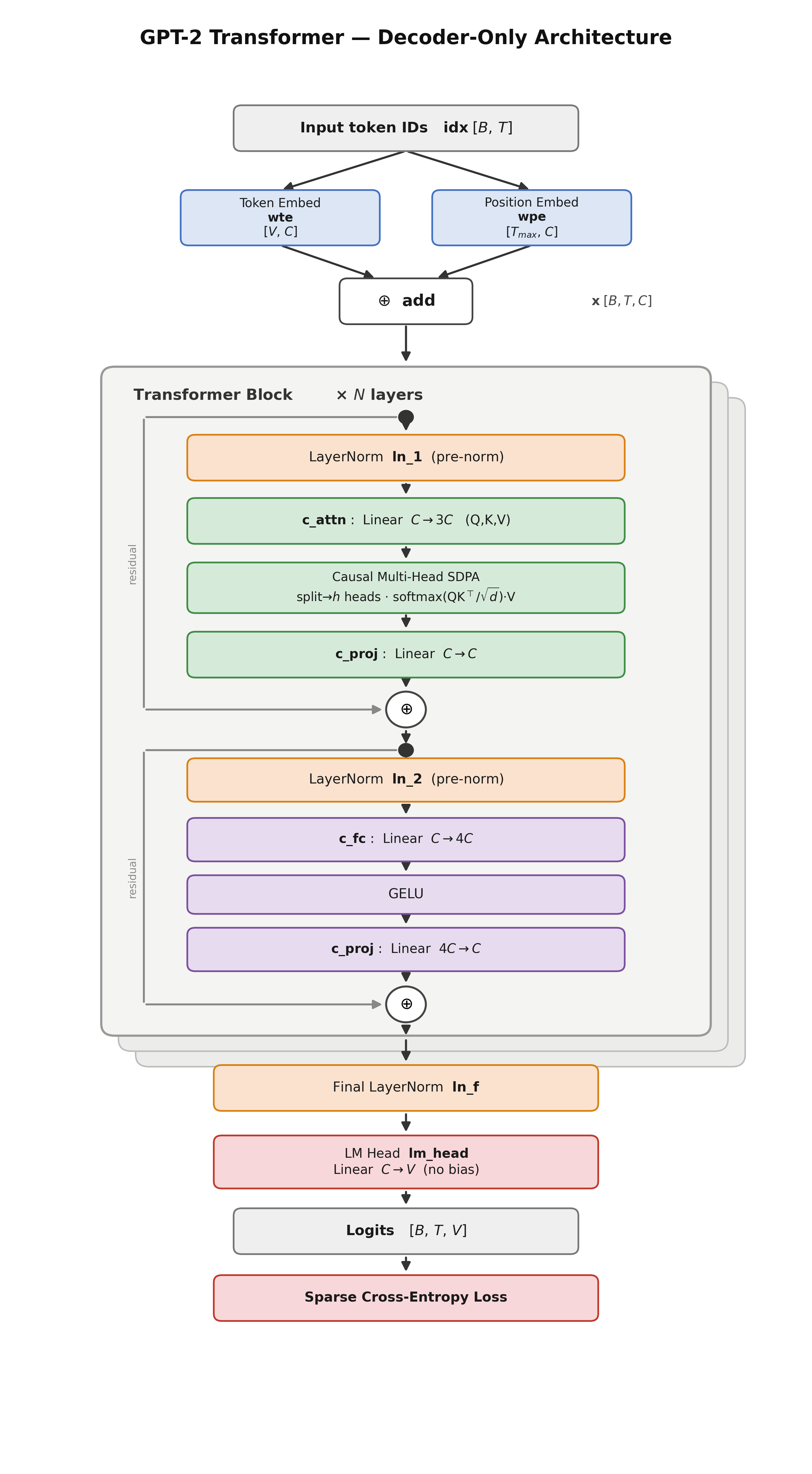}
    \caption{The GPT-2 decoder-only Transformer architecture featuring 12 pre-normalized decoder blocks, each combining masked multi-head self-attention with a GELU feed-forward sublayer through residual connections, and weight-tied embeddings.}
    \label{fig:gpt2arch}
\end{figure}

\subsubsection*{Hyperparameters}
\begin{table}[H]
\centering
\small
\begin{minipage}[t]{0.5\textwidth}
\centering
\begin{tabular}{@{}ll@{}}
\toprule
\multicolumn{2}{c}{\textbf{Model \& Training}} \\
\midrule
Model size            & 124M \\
$d_{\text{model}}$    & 768 \\
Layers                & 12 \\
Heads                 & 12 \\
Head dim              & 64 \\
FFN hidden            & 3072 \\
Non-linear activations & GELU \\
Weight tying          & On \\
Context length        & 1024 \\
Vocabulary            & 50{,}257 (50{,}304 padded) \\
Global batch          & 524{,}288 \\
Max learning rate     & $6{\times}10^{-4}$ \\
Min learning rate     & $6{\times}10^{-5}$ \\
Warmup steps          & 715 \\
Training steps        & 19{,}073 \\
Grad.\ accumulation   & 4 \\
Dataset               & 10B FineWeb-Edu~\cite{fineweb} \\
Parallelism           & DDP, 8$\times$ RTX 6000 Ada \\
\bottomrule
\end{tabular}
\end{minipage}%
\begin{minipage}[t]{0.5\textwidth}
\centering
\begin{tabular}{@{}ll@{}}
\toprule
\multicolumn{2}{c}{\textbf{Optimization \& Initialization}} \\
\midrule
Optimizer        & AdamW \\
$\beta_1$        & 0.9 \\
$\beta_2$        & 0.95 \\
$\epsilon$       & $1{\times}10^{-8}$ \\
Weight decay     & 0.1 \\
Gradient clip    & 1.0 \\
Dropout          & 0 \\
Precision        & FP32 \\
Weight init      & $\mathrm{std}=0.02$ \\
Residual proj init & $\mathrm{std}=0.02/\sqrt{2\,n_{\text{layers}}}$ \\
\bottomrule
\end{tabular}
\end{minipage}
\caption{Training hyperparameters for the 124M GPT-2 run. Residual projections (attention output and MLP down-projection) use the scaled initialization.}
\label{tab:hparams}
\end{table}

\subsubsection*{Experimental Setup}
The framework was evaluated by modelling two identical instances of the same 124M-parameter GPT-2 configuration as in Figure~\ref{fig:gpt2arch} and by training them under the same exact conditions as described in Table~\ref{tab:hparams}, only differing in the framework used to build the model. Holding architecture, initialization, data ordering, and hardware conditions identical across both isolates framework as the single independent variable. The first instance realized with BluTrain has its layer definitions, every forward and backward operator, the AdamW optimizer step~\cite{adamw}, and the parallel distribution expressed natively through the framework. The equivalent PyTorch implementation serves as the baseline. All reported latencies are derived via median CUDA-event timing across multiple kernel invocations, with numeric correctness strictly verified against a high-precision reference to ensure that hardware acceleration does not compromise mathematical fidelity.

\subsection{Numerical fidelity and convergence}
\begin{figure}[H]
    \centering
    \begin{subfigure}{0.48\textwidth}
        \centering
        \includegraphics[width=\textwidth]{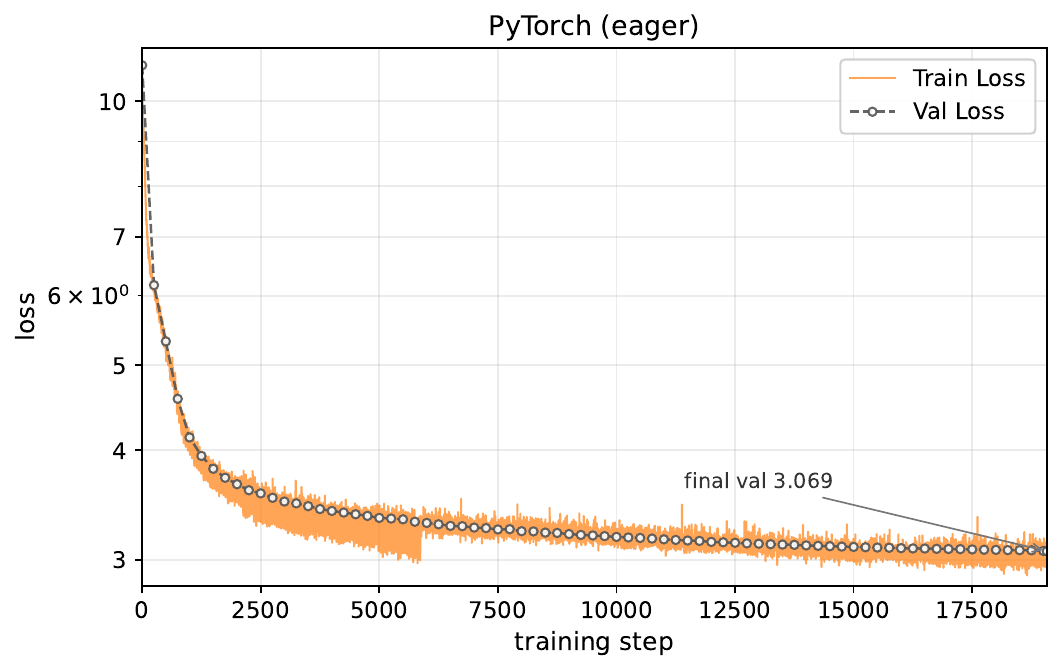}
        \caption{PyTorch (eager).}
    \end{subfigure}
    \hfill
    \begin{subfigure}{0.48\textwidth}
        \centering
        \includegraphics[width=\textwidth]{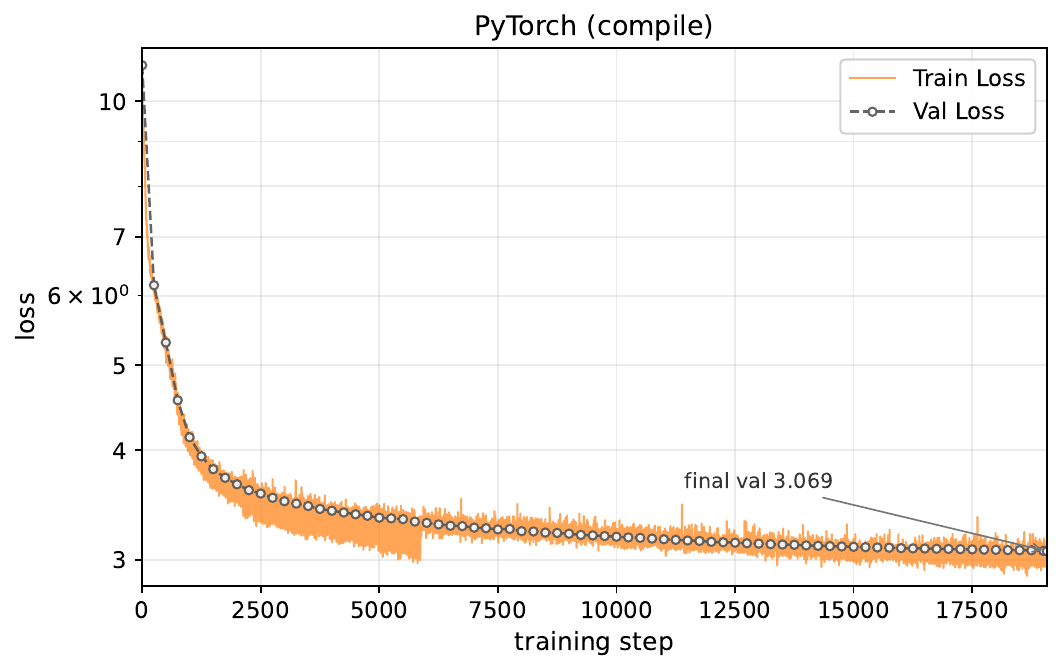}
        \caption{PyTorch (\code{compile}).}
    \end{subfigure}

    \vspace{1ex}
    \begin{subfigure}{0.48\textwidth}
        \centering
        \includegraphics[width=\textwidth]{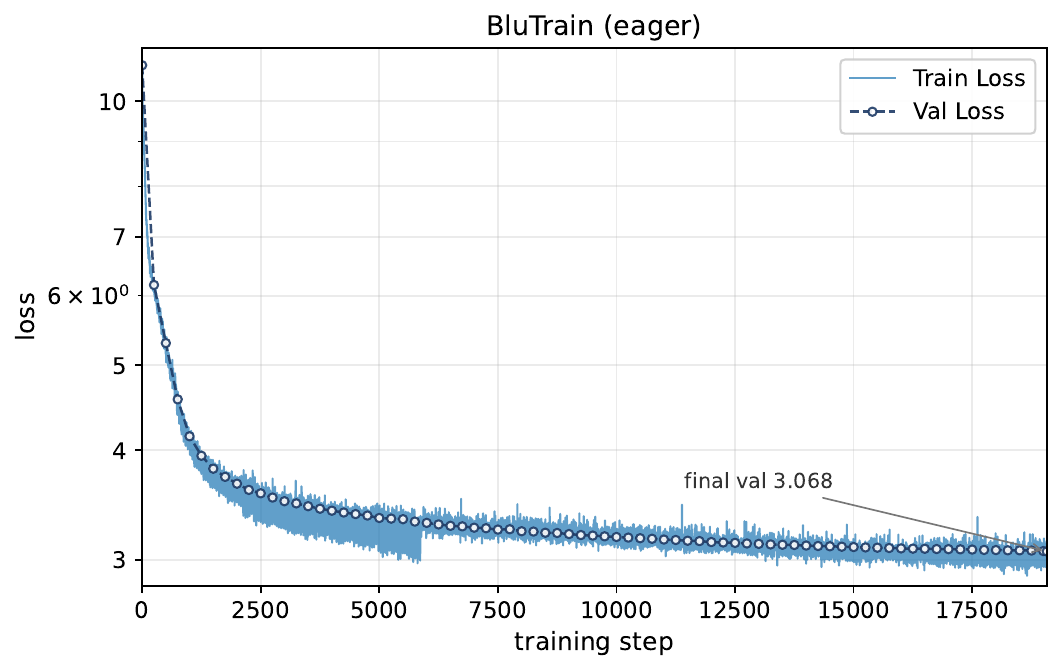}
        \caption{BluTrain (eager).}
    \end{subfigure}
    \caption{Training and validation loss over the full 19{,}073-step 124M GPT-2 run. Panels (a)--(c) show the raw per-run train/val curves. All configurations converge identically to a final validation loss of $\approx 3.07$, confirming that BluTrain preserves numerical fidelity against both PyTorch eager and \code{compile}.}
    \label{fig:valloss}
\end{figure}

The central result is that BluTrain reproduces the training trajectory rather than merely
approximating it. Figure~\ref{fig:valloss} overlays the validation-loss curves of
the two frameworks across the full run. They are visually indistinguishable: the
maximum gap between the two curves at any of the 77 logged checkpoints is below
$3\times10^{-3}$, and both descend monotonically to a final validation loss of
$\approx 3.07$ (BluTrain $3.0675$, PyTorch $3.0695$). Minimum training
loss likewise matches ($2.8771$ vs.\ $2.8793$). For a
first-principles stack this near-exact agreement is strong evidence that the
operator semantics, accumulation orders, and promotion rules are correct. This is a
direct consequence of controlling the numerics end to end.

\subsection{Throughput}
Across the identical 124M, 8-GPU, 19{,}073-step workload, BluTrain achieves an aggregate throughput of \textbf{$\approx$406{,}600 tokens/s} ($\approx 1{,}313$\,ms/step). This outperforms the PyTorch baseline of $\approx$394{,}700 tokens/s ($\approx 1{,}359$\,ms/step) by a sustained $\approx$3\% margin. In absolute terms, this implies BluTrain processes $\approx$11{,}900 more tokens every second than PyTorch during the training runtime. This end-to-end reduction in step time is the direct mathematical product of the kernel-level optimizations. In eager mode on a single GPU, BluTrain sustains $\approx$54{,}600 tokens/s, confirming efficient hardware utilization for a dense Transformer training loop at this scale.

\begin{table}[H]
\centering
\caption{End-to-end training comparison on the 124M GPT-2 run (8$\times$ RTX 6000 Ada, 19{,}073 steps, FP32). BluTrain runs in eager mode against both PyTorch eager and PyTorch \code{compile} baselines.}
\label{tab:throughput}
\small
\setlength{\tabcolsep}{8pt}
\begin{tabular}{lrrr}
\toprule
Metric & PyTorch (eager) & PyTorch (compile) & BluTrain (eager) \\
\midrule
Avg.\ throughput (tok/s) & 394{,}663 & 402{,}453 & \textbf{406{,}595} \\
Max.\ throughput (tok/s) & 419{,}302 & \textbf{431{,}097} & 430{,}925 \\
Avg.\ step time (ms)     & 1359.14   & 1326.73            & \textbf{1313.43} \\
Min.\ train loss         & 2.8793    & \textbf{2.8765}    & 2.8771 \\
Min.\ val loss           & 3.0695    & 3.0694             & \textbf{3.0675} \\
\bottomrule
\end{tabular}
\end{table}

\begin{figure}[H]
    \centering
    \includegraphics[width=\textwidth]{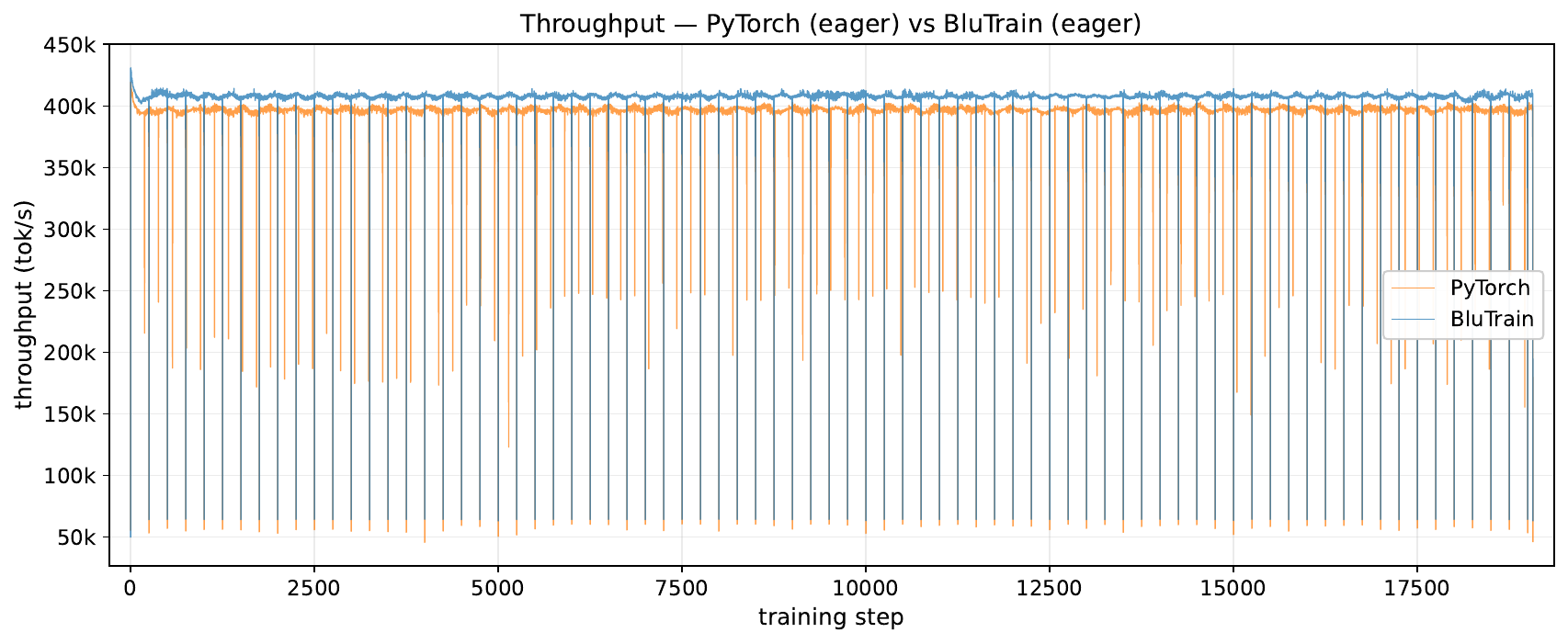}

    \vspace{1ex}
    \includegraphics[width=\textwidth]{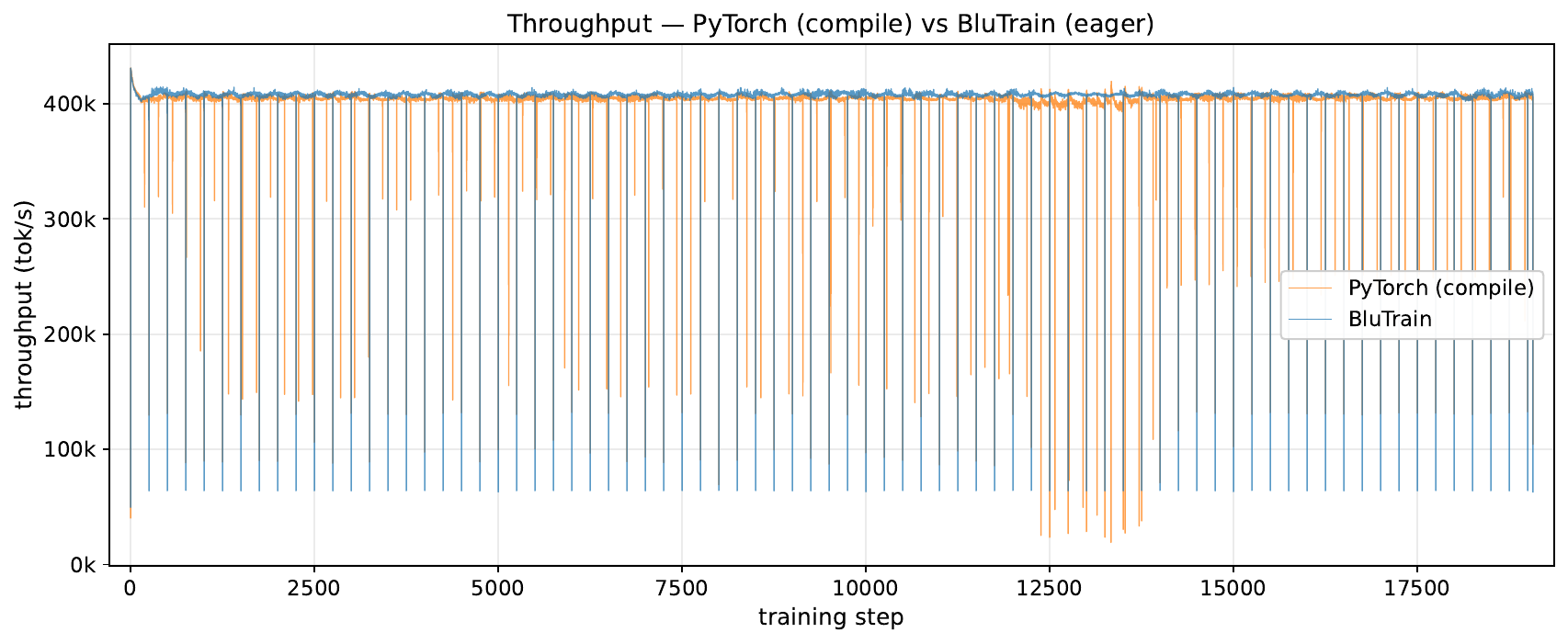}
    \caption{Per-step throughput across the full 19{,}073-step run. BluTrain in eager mode tracks at or above both the PyTorch eager and \code{compile} baselines; the periodic dips coincide with checkpoint and evaluation steps.}
    \label{fig:throughput_curves}
\end{figure}

\subsection{Memory}
As demonstrated in Table~\ref{tab:mem}, BluTrain consistently operates with a smaller memory footprint than PyTorch. During steady-state execution, BluTrain stabilizes at \textbf{21.48\,GiB} per GPU, compared to PyTorch's 27.54\,GiB (default), 26.35\,GiB (power-of-two allocator), and 24.38\,GiB (compile mode), representing a \textbf{22\% reduction} in sustained VRAM for the given configuration. This footprint reduction is obtained directly from BluTrain's deterministic caching allocator through two primary mechanisms. First, the allocator strictly minimizes internal fragmentation by dynamically tuning tensor alignment based on allocation frequencies and historical wasted bytes. Second, by automatically executing a targeted cache flush following the anomalous 0th step (which incorporates validation and generation passes), BluTrain permanently returns 5--7\% of allocated memory to the driver. When this identical cache-flush strategy is tested on PyTorch, its size-class heuristics immediately reacquire the released memory, resulting in a 1--3\% footprint spike rather than a reduction. This fundamental allocator difference yields strictly more VRAM headroom for scaling batch sizes on equivalent hardware.

\begin{table}[H]
\centering
\caption{GPU memory (GiB) during the 124M run. Configuration: $[B, T, C] = [16, 1024, 768]$. ``Train'' is steady state.}
\label{tab:mem}
\small
\begin{tabular}{lrrr}
\toprule
Framework & Step 0 & Step 1 & Train \\
\midrule
BluTrain                 & \textbf{23.01} & \textbf{23.01} & \textbf{21.48} \\
PyTorch                  & 24.09 & 27.16 & 27.54 \\
PyTorch (pow-2 divisions)& 28.04 & 29.05 & 26.35 \\
PyTorch (Compile Mode)   & 20.22 & 23.29 & 24.38 \\
\bottomrule
\end{tabular}
\end{table}

\subsection{Long-context training}
\label{subsec:longctx}
To leverage long-context training, which is bounded by memory rather than by compute, the context length in the existing 124M GPT-2 configuration was increased from 1024 to 16{,}384 while the batch size was reduced from 16 to 2, as listed in Table~\ref{tab:longctx}. On a single RTX 6000 Ada (48\,GiB), BluTrain reaches a peak footprint of 40.4\,GiB against PyTorch's 47.8\,GiB, a 15\% footprint reduction at the identical model, batch, sequence length, and precision, and it does so at a higher throughput of 17{,}784 versus 14{,}849 tokens/s. PyTorch runs at 99.6\% occupancy with roughly 0.2\,GiB of headroom, leaving it one allocator-fragmentation spike away from an out-of-memory failure, whereas BluTrain retains a 7.6\,GiB reserve on the same card while running faster. It converts directly into reachable sequence length. Modelling the same configuration would demand recomputations at throughput cost, smaller micro-batches that add gradient-accumulation steps and overhead, or sharding the model across multiple GPUs. BluTrain instead fits natively on one device and keeps full single-GPU throughput. Training at 99.6\% of VRAM is fragile, turning the training prone to mid-run out-of-memory crashes caused by allocator fragmentation and transient spikes, and a multi-gibibyte reserve keeps runs stable.

\begin{table}[H]
\centering
\caption{Long-context training of the 124M GPT-2 on a single RTX 6000 Ada (48\,GiB), FP32. Both frameworks use 2 micro-batches at a context length of 16{,}384.}
\label{tab:longctx}
\small
\setlength{\tabcolsep}{8pt}
\begin{tabular}{lrr}
\toprule
Metric & PyTorch & BluTrain \\
\midrule
Micro-batches      & 2           & 2 \\
Context length     & 16{,}384    & 16{,}384 \\
Peak memory (GiB)  & 47.8        & \textbf{40.4} \\
Headroom (GiB)     & 0.2 (0.4\%) & \textbf{7.6 (15.8\%)} \\
Throughput (tok/s) & 14{,}849    & \textbf{17{,}784} \\
\bottomrule
\end{tabular}
\end{table}

\subsection{Largest trainable parameter model}
\label{subsec:maxparam}
The configuration in Table~\ref{tab:maxparam_cfg} specifies a 2.42-billion-parameter GPT-2-style model. On a single RTX 6000 Ada (48\,GiB), neither PyTorch eager mode nor PyTorch compile mode can fit and train this model, both terminating with an out-of-memory failure. As reported in Table~\ref{tab:maxparam_res}, BluTrain trains the identical model on the same single chip in eager mode, reaching a peak footprint of 46.9\,GiB and sustaining a throughput of 13.6K tokens/s. This establishes a larger trainable-parameter ceiling on fixed hardware. The same 48\,GiB device that cannot admit the model under PyTorch trains it natively under BluTrain, without recourse to model parallelism, offloading, or activation checkpointing.

\begin{table}[H]
\centering
\caption{Model configuration for the largest-trainable-model experiment, a 2.42\,B-parameter GPT-2 configuration.}
\label{tab:maxparam_cfg}
\small
\begin{tabular}{@{}ll@{}}
\toprule
\multicolumn{2}{c}{\textbf{2.42\,B-parameter GPT-2}} \\
\midrule
Batch size $B$       & 2 \\
Context length $T$   & 1024 \\
$n_\text{embd}$      & 3072 \\
$n_\text{layers}$    & 20 \\
$n_\text{heads}$     & 24 \\
Head dim             & 128 \\
Global batch         & 524{,}288 \\
\bottomrule
\end{tabular}
\end{table}

\begin{table}[H]
\centering
\caption{Peak memory and throughput on a single RTX 6000 Ada (48\,GiB), FP32. PyTorch fails to fit the model in either eager or compile mode, while BluTrain trains it natively.}
\label{tab:maxparam_res}
\small
\setlength{\tabcolsep}{8pt}
\begin{tabular}{lccc}
\toprule
Metric & PyTorch (eager) & PyTorch (compile) & BluTrain (eager) \\
\midrule
Peak memory (GiB)  & OOM & OOM & \textbf{46.9} \\
Throughput (tok/s) & --- & --- & \textbf{13{,}600} \\
\bottomrule
\end{tabular}
\end{table}


\section{Related Work}
\label{sec:related}

BluTrain sits at the intersection of deep-learning programming models, distributed-execution systems, and low-level kernel optimization. Its architectural philosophy mirrors the custom, internal execution stacks deployed at frontier AI research labs, rather than generic open-source ecosystems.

\subsubsection*{Deep Learning Frameworks}
General-purpose frameworks such as PyTorch~\cite{pytorch,pytorch2} and TensorFlow~\cite{tensorflow} prioritize user flexibility and rapid prototyping. To achieve this, they rely heavily on dynamic computational graphs, Python runtimes, and JIT tracing mechanisms (e.g., TorchDynamo~\cite{torchdynamo}). While versatile, these abstractions inherently decouple the user-facing tensor logic from the backend hardware. Even though the underlying kernels execute in C++ or CUDA, the host-side Python interpreter must continuously acquire the Global Interpreter Lock (GIL) to dispatch operations and manage object reference counts. For high-speed GPU execution, this dispatch cycle often becomes a bottleneck, causing the GPU to starve while waiting for Python to issue the next command. In contrast, BluTrain explicitly eschews Python-level flexibility in favor of compile-time specialization. By operating as a fully native, statically linked C++ runtime, BluTrain completely bypasses these host-side bottlenecks to enforce absolute execution determinism.

\subsubsection*{Distributed Scale and Parallelism} Distributing the training of large neural networks has yielded a family of complementary parallelism strategies, each addressing a distinct resource bottleneck. Data parallelism is the most established: Horovod~\cite{horovod} and PyTorch DistributedDataParallel~\cite{pytorch_ddp} replicate the model across workers and synchronize gradients through all-reduce, with the latter overlapping communication with the backward pass via bucketed gradient reductions. Because full replication leaves model state larger than device memory, sharded data-parallel approaches (ZeRO/DeepSpeed~\cite{zero,zero_offload,zero_infinity,deepspeed} and FSDP~\cite{fsdp}) instead partition optimizer states, gradients, and parameters across data-parallel ranks. Orthogonally, tensor (intra-layer) parallelism, introduced in Mesh-TensorFlow~\cite{meshtensorflow} and Megatron-LM~\cite{megatron}, splits individual weight matrices across devices to relieve per-layer memory and compute. The growth of context windows has more recently motivated context (sequence) parallelism, which shards the sequence dimension; Megatron sequence parallelism, Ring Attention~\cite{ringattn}, and DeepSpeed-Ulysses~\cite{ulysses} differ chiefly in how activations are exchanged to compute attention over a partitioned sequence. A substantial body of work further composes these axes into combined multi-dimensional schemes for extreme scale, typically tightly coupled to a particular model implementation. In contrast, DTMS exposes data, tensor, and context parallelism as independent strategies decoupled from the model definition, so that each can be selected according to the workload's bottleneck and the underlying network topology.

\subsubsection*{Deep Learning Compilers and Memory Optimization} To circumvent the overhead inherent to high-level frameworks, the industry has heavily invested in Deep Learning (DL) compilers such as Triton~\cite{triton,triton2}, TVM~\cite{tvm}, and XLA~\cite{xla}. These systems dynamically trace execution graphs to perform Just-In-Time (JIT) operator fusion, attempting to recover the memory bandwidth lost by fragmented kernel launches. Concurrently, memory management has become a critical bottleneck~\cite{checkpointing,capuchin}; static graph compiler frequently induces severe memory fragmentation, forcing frameworks to rely on complex, heuristic-based caching allocators. BluTrain addresses both compute and memory from a static, holistic perspective. Computations are globally optimized and structurally fused via a native MLIR-based deep-learning compiler. Memory is actively managed by a deterministic, self-tuning block-pool allocator engineered specifically to exploit the highly predictable, cyclic memory lifecycles (forward, backward, optimize) of neural network training. By operating with fixed, pre-allocated memory arenas rather than generic size-class heuristics, BluTrain structurally eliminates allocation fragmentation at the root.

\subsubsection*{Kernel-Level Optimization and Microarchitecture} At the absolute limit of hardware performance, maximizing Model FLOPs Utilization (MFU) demands extreme register and shared-memory locality across every computational phase. To achieve this, the industry relies heavily on generalized, black-box libraries (e.g., cuDNN, cuBLAS) or broad template frameworks (e.g., CUTLASS~\cite{cutlass}) to execute everything from attention mechanisms to optimization steps. While these generalized toolkits offer broad hardware portability and rapid development, they inherently compromise fine-grained microarchitectural control. BluTrain explicitly rejects these intermediate abstractions across its entire stack. By engineering every operation, from the dense linear algebra engines down to the normalization and optimizer kernels, directly in raw CUDA and native PTX, the framework secures uncompromising, cycle-accurate authority over the hardware. This holistic bypass enables precise warp-level instruction issue, deterministic sub-byte accumulation, and numerically stable online reductions~\cite{onlinesoftmax,welford} across the entire model pipeline, extracting the absolute maximum mathematical fidelity directly from the silicon.

\section{Discussion and Limitations}
\label{sec:discussion}

We are deliberate about what the evidence does and does not show.

\emph{Throughput is an honest, foundational win.} BluTrain is $\approx 3\%$ faster end-to-end ($\sim$407K vs.\ $\sim$395K tokens/s): the per-kernel wins on attention and loss do not compound into a larger margin because the remaining kernels are at parity by construction (\sectref{sec:kernels}). However, this robust few-percent advantage serves strictly as an initial baseline. The system removes the rigid overheads of high-level framework wrappers. The structural foundation is now validated; with direct control over every computational layer, the ultimate performance ceiling is the framework's own to raise.

\emph{Validation is currently at GPT-2 scale.} The design makes no
Transformer-specific assumptions, but the end-to-end evidence here is from one
model family; broader validation across architectures is future work
(\sectref{sec:conclusion}).

\emph{Lack of systematic component ablations.} While the current evaluation demonstrates the compounding end-to-end benefits of the integrated architecture, we do not currently provide a quantitative breakdown isolating the exact performance contribution of each individual subsystem. Rigorous, component-level ablations to explicitly quantify the isolated impact of every microarchitectural design decision remain pending.

\emph{Orchestration and failure handling limitations.} The fault taxonomy, while systematically comprehensive in classification, reflects principled design reasoning more than accumulated operational experience. Recovery logic for several fault classes awaits validation against the irregular, compounding failure signatures that emerge from heterogeneous GPU populations, aging silicon, and adverse datacenter conditions. Predictive fault detection thresholds likewise remain to be calibrated against per-device empirical baselines.

\emph{Distributed scalability limitations.} Current validation is scoped to a single 8-GPU node, where intercommunication pressure and rank-count-dependent contention are negligible. Characterizing the DTMS stack under multi-node topologies (where network contention, asymmetric link degradation, and correlated multi-rank failures become dominant operational realities) remains an important next step toward establishing the distributed architecture's production readiness at scale.

\section{Conclusion and Future Work}
\label{sec:conclusion}

In this report, we detailed the architectural foundations of BluTrain, validating its structural efficiency and native execution capabilities. While ongoing optimizations will continue to refine the system's distributed scalability and computational efficiency, the underlying framework is now securely established. Furthermore, while the current runtime is strictly specialized for NVIDIA microarchitectures, future iterations will evolve the deep-learning compiler into a fully hardware-agnostic backend. This will decouple the core execution engine from specific physical constraints, extending BluTrain's deterministic performance optimizations across diverse accelerator ecosystems.

Concurrently, we are conducting a systematic investigation into the microarchitectural factors governing numerical precision. By empirically evaluating how varying combinations of arithmetic execution formats, fused operation ordering, and low-level accumulation strategies impact end-to-end training and validation loss, we aim to establish a fully deterministic model of hardware-level mathematical fidelity. This ongoing research will ensure that future high-throughput scaling efforts continue to yield superior convergence trajectories.

Driven by our foundational mission in applied AI research, our overarching vision is to leverage deep learning to solve complex, real-world problems. With the structural efficiency of BluTrain established, our focus shifts toward training highly competitive, large-scale models across diverse domains. We intend to expand our research across all primary modalities: Natural Language Processing (NLP), Computer Vision, Speech, Recommender Systems, and advanced Generative Multimodal architectures. Possessing this native computational foundation equips us with the structural authority to push the boundaries of AI research and tackle these real-world challenges without compromise.


\clearpage
\appendix

\begin{center}
    {\LARGE\bfseries Appendix}
\end{center}
\vspace{1.5em}

\section{Hardware and Benchmarking Methodology}
\label{app:hwmethod}
All empirical kernel benchmarks and performance metrics presented in Appendix~\ref{app:attn_fwd} through~\ref{app:adam} were executed on a single NVIDIA RTX 6000 Ada Lovelace (48\,GB) GPU. (Hardware specifications for distributed scaling benchmarks are provided locally within their respective sections). To ensure rigorous, noise-free telemetry, the benchmarking methodology adheres to the following constraints:
\begin{itemize}\itemsep2pt
    \item \textbf{Warm-up Initialization:} 25 unrecorded iterations per kernel are executed to stabilize SM clock frequencies.
    \item \textbf{Iteration Count:} 100 timed runs are executed per configuration, reporting the median latency.
    \item \textbf{Cache Eviction:} The L2 cache is explicitly flushed between iterations to prevent hot-cache bias and artificially inflated bandwidth metrics.
    \item \textbf{Telemetry:} Latency is measured strictly via asynchronous CUDA events, isolating true GPU execution time from host-side dispatch overhead.
\end{itemize}

\section{Performance Benchmarks}
\label{app:benchmarks}
The following subsections detail the absolute performance metrics (Latency, TFLOPS, and Memory Bandwidth) of individual kernels across varying tensor shapes. The empirical data is extracted directly from Nsight Compute profiles and hardware telemetry, ensuring exact reproduction of the execution state without manipulation.

\subsection{Attention Forward \small (\textcolor{bluteal}{\rule{1.2ex}{1.2ex}} BluBridge, \textcolor{ptrose}{\rule{1.2ex}{1.2ex}} PyTorch)}
\label{app:attn_fwd}

\begin{figure}[H]
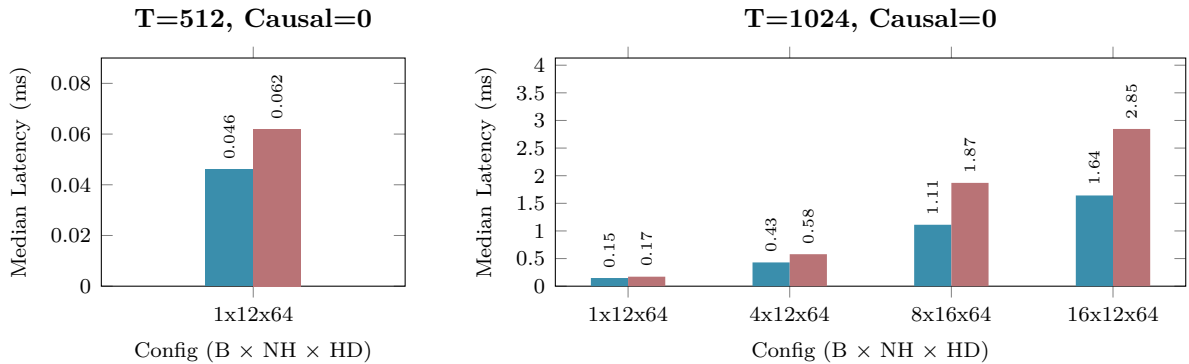

    \centering
    \begin{subfigure}{0.35\textwidth}
        \centering

    \end{subfigure}
    \caption{Forward Attention median latency (Causal=0). Lower is better.}
\end{figure}

\begin{figure}[H]
    \centering
    \begin{subfigure}{0.35\textwidth}
        \centering
        %
    \end{subfigure}
    \caption{Forward Attention median latency (Causal=1). Lower is better.}
\end{figure}

\begin{figure}[H]
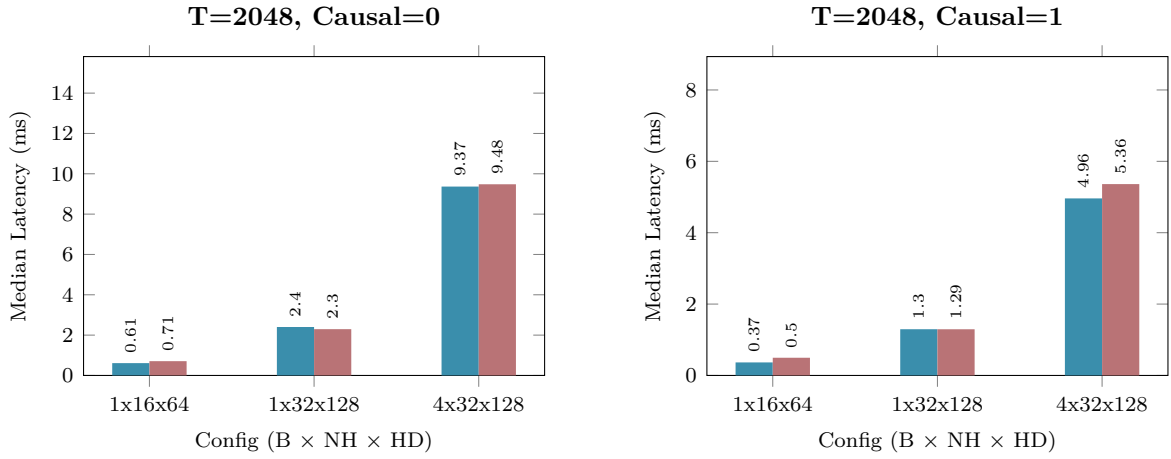

    \centering
    \begin{subfigure}{0.48\textwidth}
        \centering
        %
    \end{subfigure}
    \caption{Forward Attention median latency (T=2048 configs). Lower is better.}
\end{figure}

\begin{figure}[H]
    \centering
    \begin{subfigure}{0.35\textwidth}
        \centering
        %
    \end{subfigure}
    \caption{Forward Attention TFLOPS (Causal=0). Higher is better.}
\end{figure}

\begin{figure}[H]
    \centering
    \begin{subfigure}{0.35\textwidth}
        \centering
        %
    \end{subfigure}
    \caption{Forward Attention TFLOPS (Causal=1). Higher is better.}
\end{figure}

\begin{figure}[H]
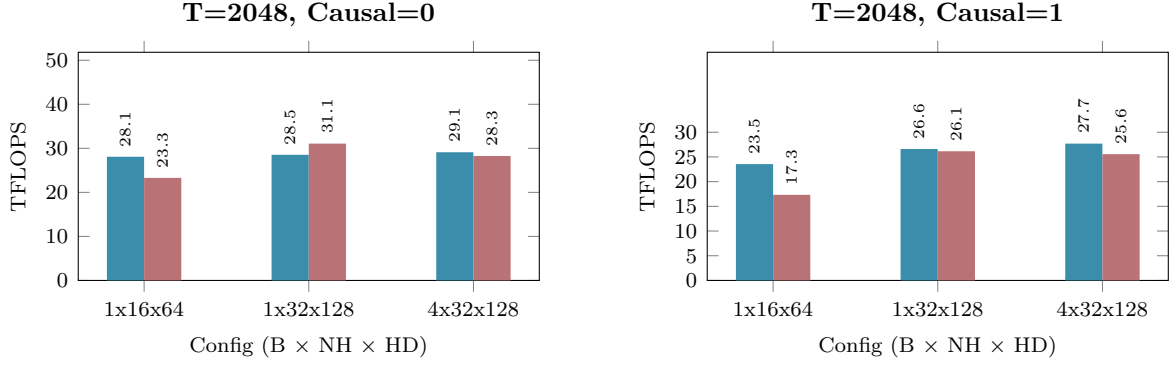

    \centering
    \begin{subfigure}{0.48\textwidth}
        \centering
        %
    \end{subfigure}
    \caption{Forward Attention TFLOPS (T=2048 configs). Higher is better.}
\end{figure}

\subsection{Attention Backward \small (\textcolor{bluteal}{\rule{1.2ex}{1.2ex}} BluBridge, \textcolor{ptrose}{\rule{1.2ex}{1.2ex}} PyTorch)}
\label{app:attn_bwd}

\begin{figure}[H]
    \centering
    \begin{subfigure}{0.35\textwidth}
        \centering
        %
    \end{subfigure}
    \caption{Backward Attention median latency (Causal=0). Lower is better.}
\end{figure}

\begin{figure}[H]
    \centering
    \begin{subfigure}{0.35\textwidth}
        \centering
        %
    \end{subfigure}
    \caption{Backward Attention median latency (Causal=1). Lower is better.}
\end{figure}

\begin{figure}[H]
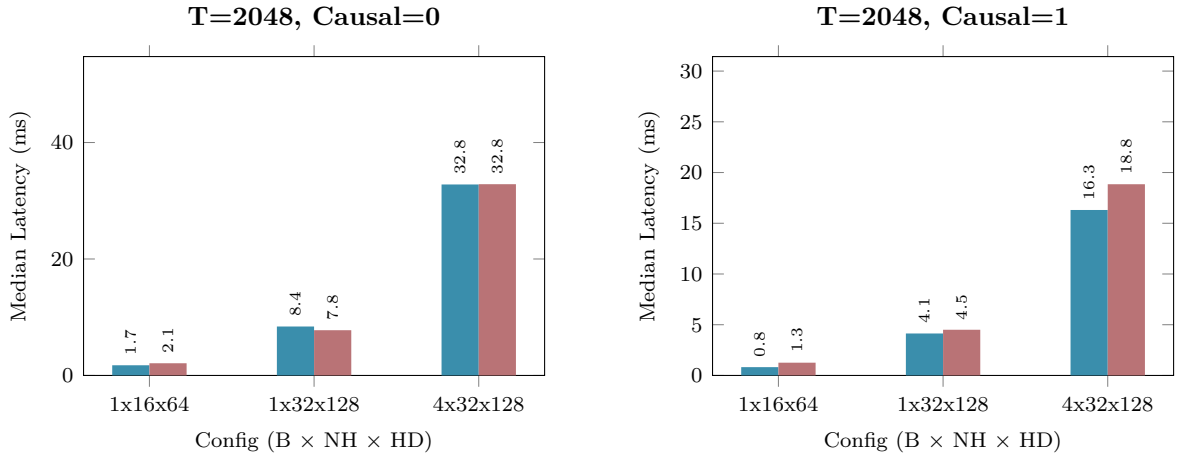

    \centering
    \begin{subfigure}{0.48\textwidth}
        \centering
        %
    \end{subfigure}
    \caption{Backward Attention median latency (T=2048 configs). Lower is better.}
\end{figure}

\begin{figure}[H]
    \centering
    \begin{subfigure}{0.35\textwidth}
        \centering
        %
    \end{subfigure}
    \caption{Backward Attention TFLOPS (Causal=0). Higher is better.}
\end{figure}

\begin{figure}[H]
    \centering
    \begin{subfigure}{0.35\textwidth}
        \centering
        %
    \end{subfigure}
    \caption{Backward Attention TFLOPS (Causal=1). Higher is better.}
\end{figure}

\begin{figure}[H]
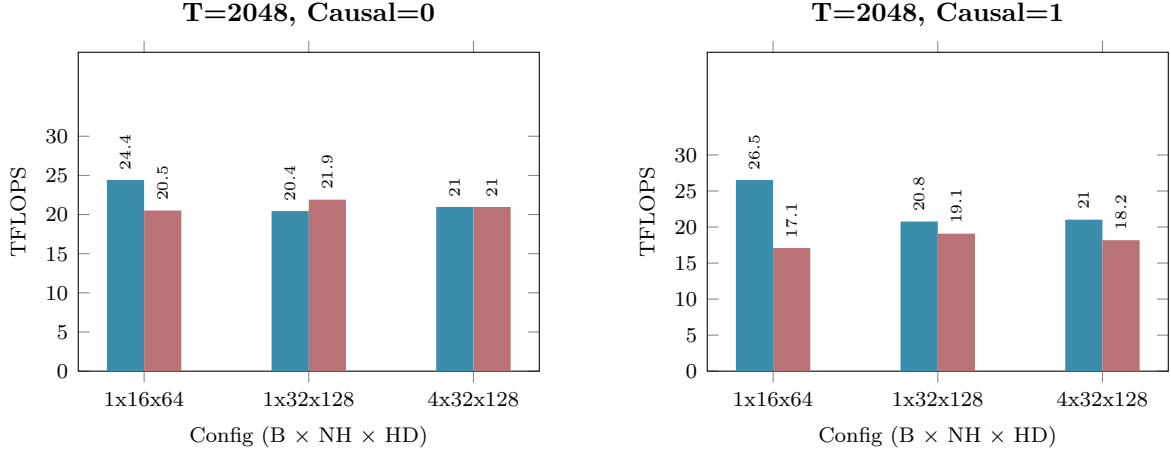

    \centering
    \begin{subfigure}{0.48\textwidth}
        \centering
        %
    \end{subfigure}
    \caption{Backward Attention TFLOPS (T=2048 configs). Higher is better.}
\end{figure}

\subsection{GELU Forward \small (\textcolor{bluteal}{\rule{1.2ex}{1.2ex}} BluBridge, \textcolor{ptrose}{\rule{1.2ex}{1.2ex}} PyTorch)}
\label{app:gelu_fwd}

\begin{figure}[H]
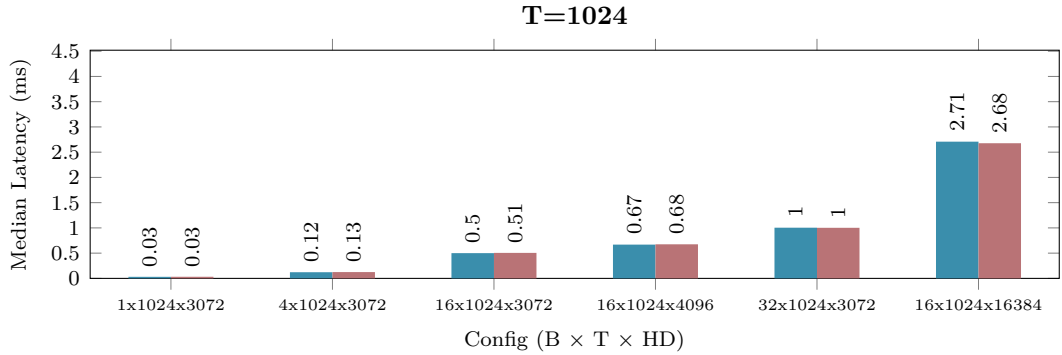

    \centering
    \begin{subfigure}{0.9\textwidth}
        \centering
        %
    \end{subfigure}
    \caption{Forward GELU median latency (T=1024). Lower is better.}
\end{figure}

\begin{figure}[H]
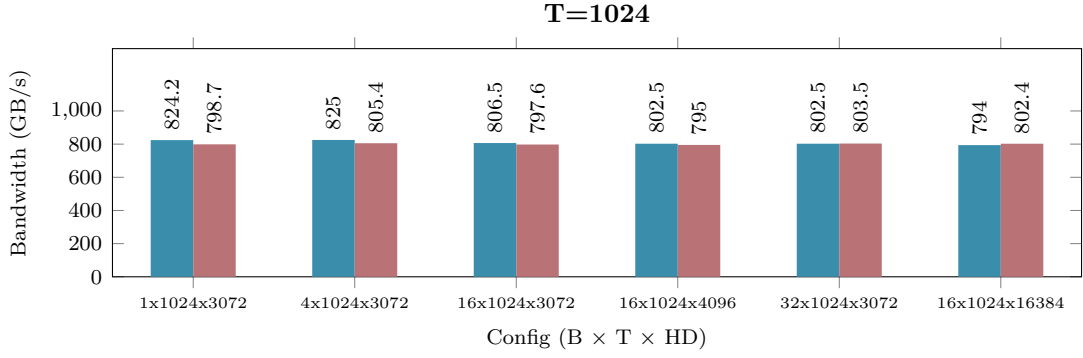

    \centering
    \begin{subfigure}{0.9\textwidth}
        \centering
        %
    \end{subfigure}
    \caption{Forward GELU memory bandwidth (T=1024). Higher is better.}
\end{figure}

\begin{figure}[H]
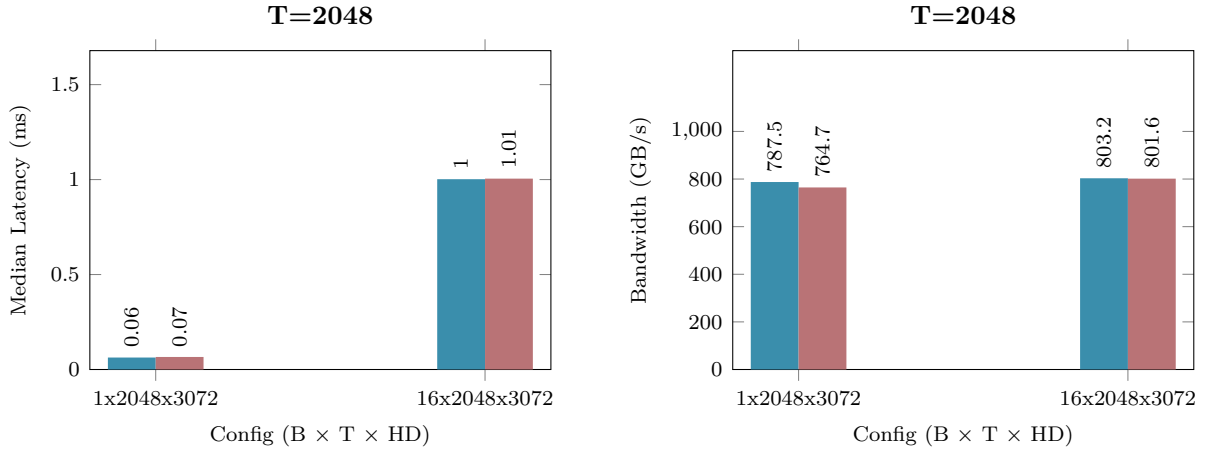

    \centering
    \begin{subfigure}{0.48\textwidth}
        \centering
        %
    \end{subfigure}
    \caption{Forward GELU performance (T=2048). Left: Median Latency (Lower is better). Right: Memory Bandwidth (Higher is better).}
\end{figure}

\subsection{GELU Backward \small (\textcolor{bluteal}{\rule{1.2ex}{1.2ex}} BluBridge, \textcolor{ptrose}{\rule{1.2ex}{1.2ex}} PyTorch)}
\label{app:gelu_bwd}

\begin{figure}[H]
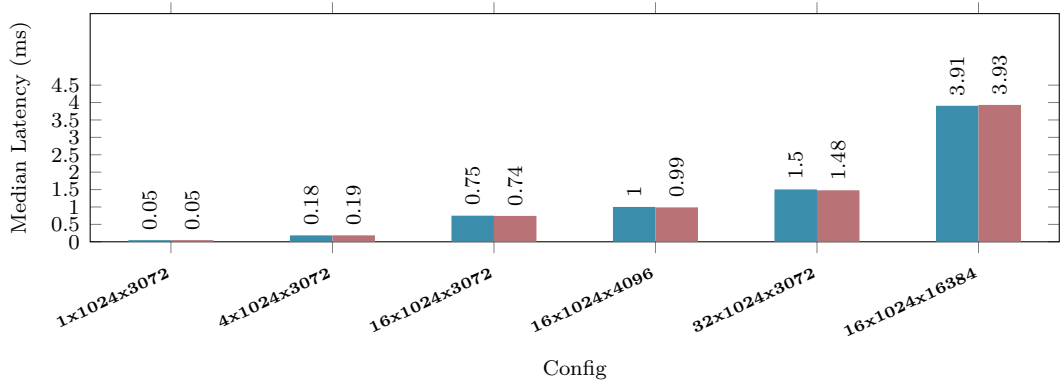

    \centering
    \begin{subfigure}{0.9\textwidth}
        \centering
        %
    \end{subfigure}
    \caption{Backward GELU median latency (T=1024). Lower is better.}
\end{figure}

\begin{figure}[H]
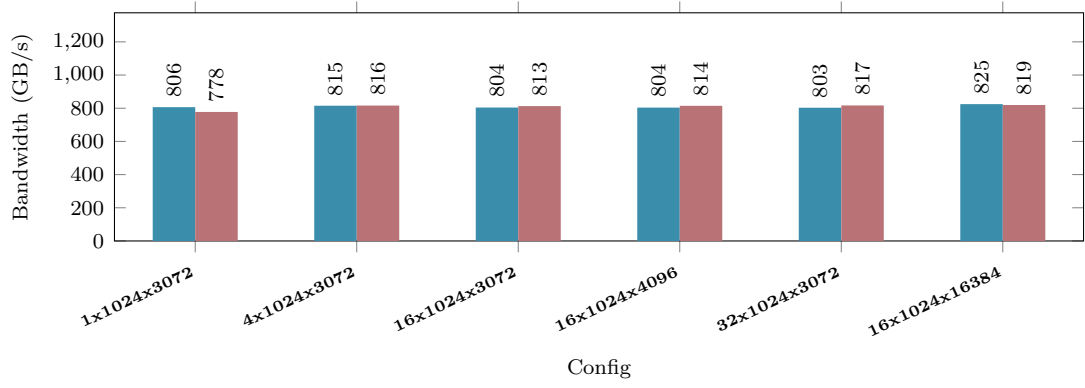

    \centering
    \begin{subfigure}{0.9\textwidth}
        \centering
        %
    \end{subfigure}
    \caption{Backward GELU memory bandwidth (T=1024). Higher is better.}
\end{figure}

\begin{figure}[H]
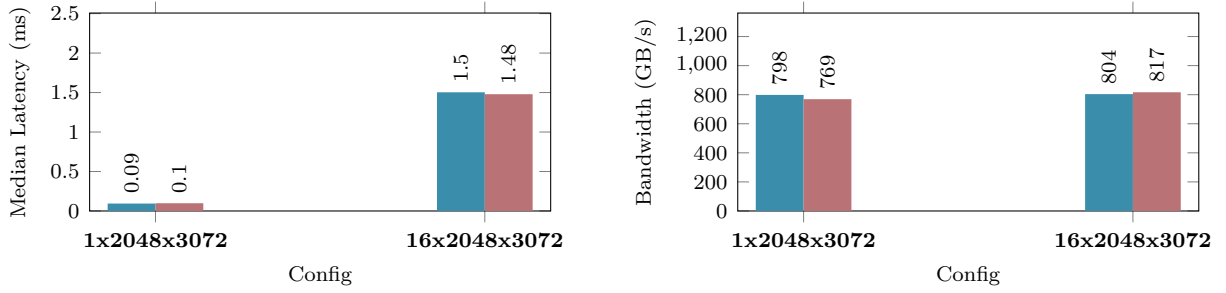

    \centering
    \begin{subfigure}{0.48\textwidth}
        \centering
        %
    \end{subfigure}
    \caption{Backward GELU performance (T=2048). Left: Median Latency. Right: Bandwidth.}
\end{figure}

\subsection{LayerNorm Forward \small (\textcolor{bluteal}{\rule{1.2ex}{1.2ex}} BluBridge, \textcolor{ptrose}{\rule{1.2ex}{1.2ex}} PyTorch)}
\label{app:ln_fwd}

\begin{figure}[H]
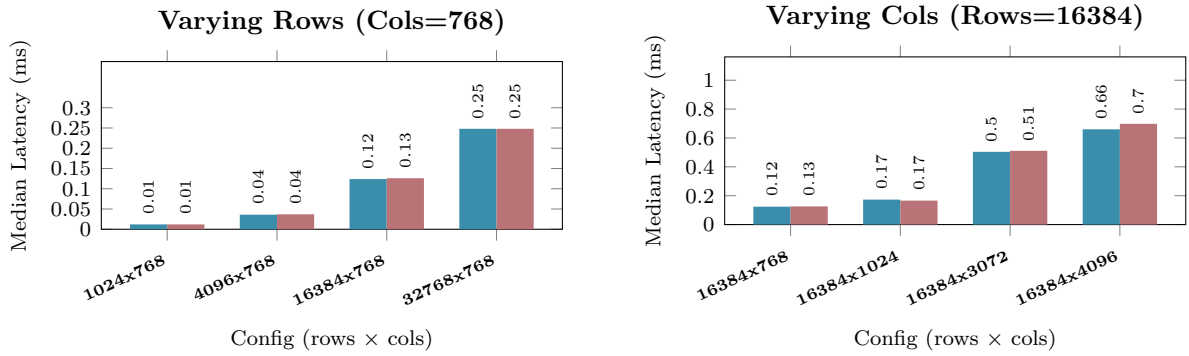

    \centering
    \begin{subfigure}{0.48\textwidth}
        \centering
        %
    \end{subfigure}
    \caption{Forward LayerNorm median latency. Lower is better.}
\end{figure}

\begin{figure}[H]
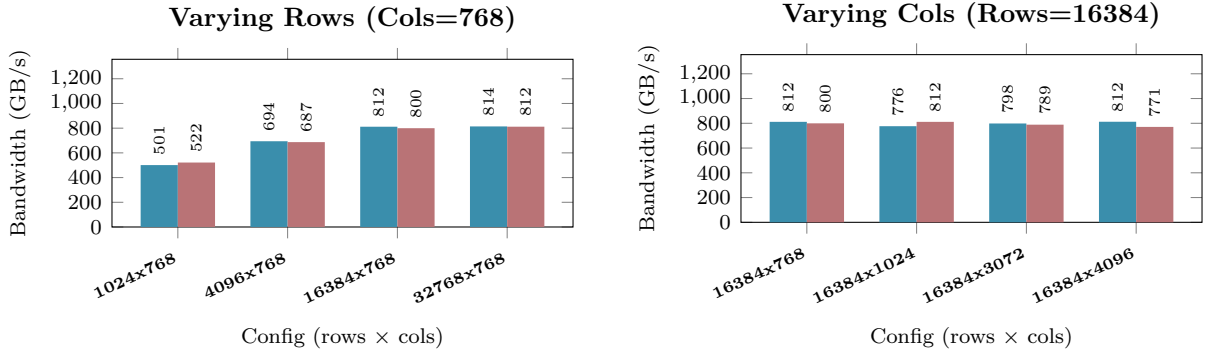

    \centering
    \begin{subfigure}{0.48\textwidth}
        \centering
        %
    \end{subfigure}
    \caption{Forward LayerNorm memory bandwidth. Higher is better.}
\end{figure}

\subsection{LayerNorm Backward \small (\textcolor{bluteal}{\rule{1.2ex}{1.2ex}} BluBridge, \textcolor{ptrose}{\rule{1.2ex}{1.2ex}} PyTorch)}
\label{app:ln_bwd}

\begin{figure}[H]
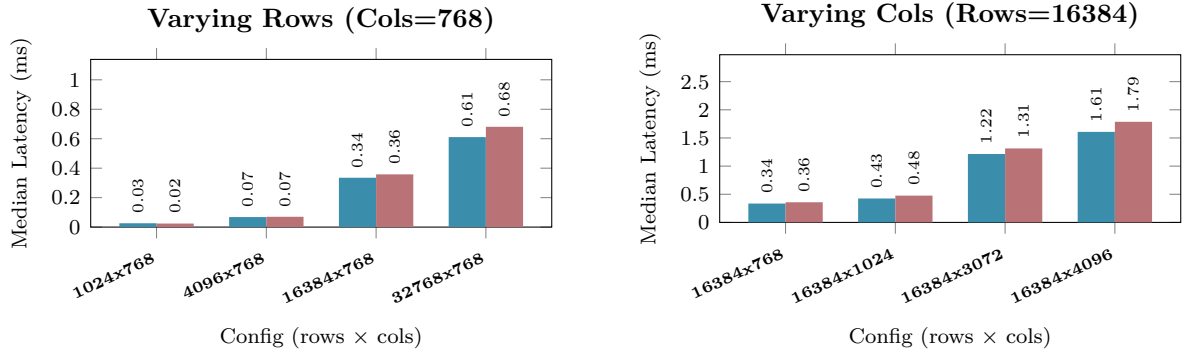

    \centering
    \begin{subfigure}{0.48\textwidth}
        \centering
        %
    \end{subfigure}
    \caption{Backward LayerNorm median latency. Lower is better.}
\end{figure}

\begin{figure}[H]
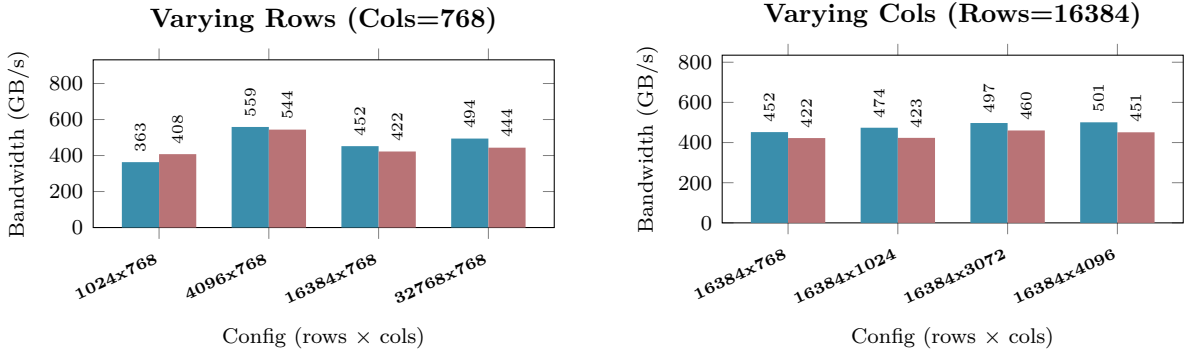

    \centering
    \begin{subfigure}{0.48\textwidth}
        \centering
        %
    \end{subfigure}
    \caption{Backward LayerNorm memory bandwidth. Higher is better.}
\end{figure}

\subsection{Sparse Cross-Entropy Loss Forward \small (\textcolor{bluteal}{\rule{1.2ex}{1.2ex}} BluBridge, \textcolor{ptrose}{\rule{1.2ex}{1.2ex}} PyTorch)}
\label{app:loss_fwd}

\begin{figure}[H]
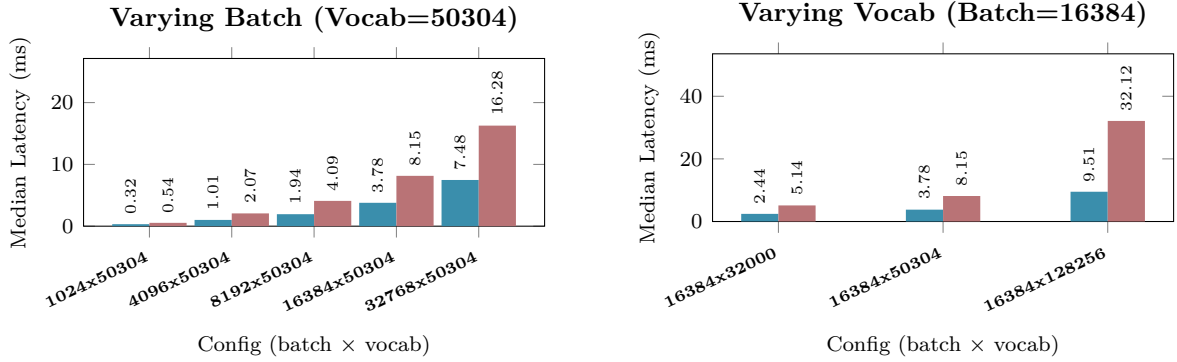

    \centering
    \begin{subfigure}{0.48\textwidth}
        \centering
        %
    \end{subfigure}
    \caption{Forward Loss median latency. Lower is better.}
\end{figure}

\begin{figure}[H]
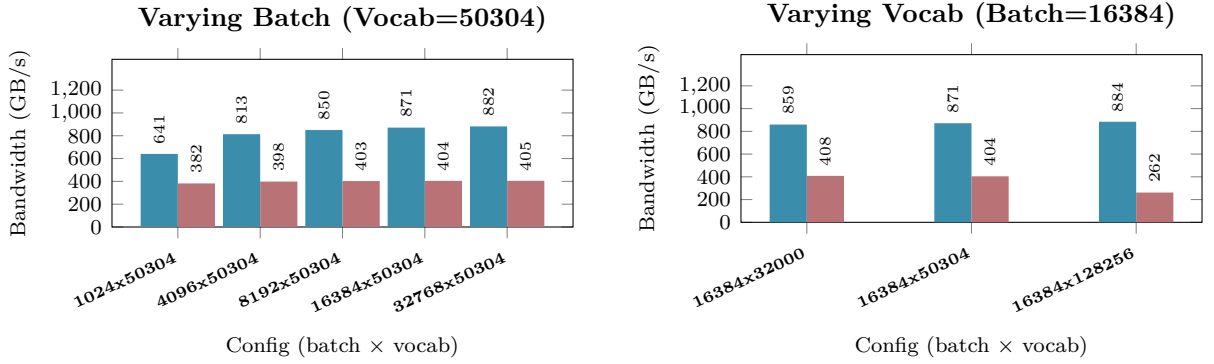

    \centering
    \begin{subfigure}{0.48\textwidth}
        \centering
        %
    \end{subfigure}
    \caption{Forward Loss memory bandwidth. Higher is better.}
\end{figure}

\subsection{Sparse Cross-Entropy Loss Backward \small (\textcolor{bluteal}{\rule{1.2ex}{1.2ex}} BluBridge, \textcolor{ptrose}{\rule{1.2ex}{1.2ex}} PyTorch)}
\label{app:loss_bwd}

\vspace*{\fill}
\begin{figure}[H]
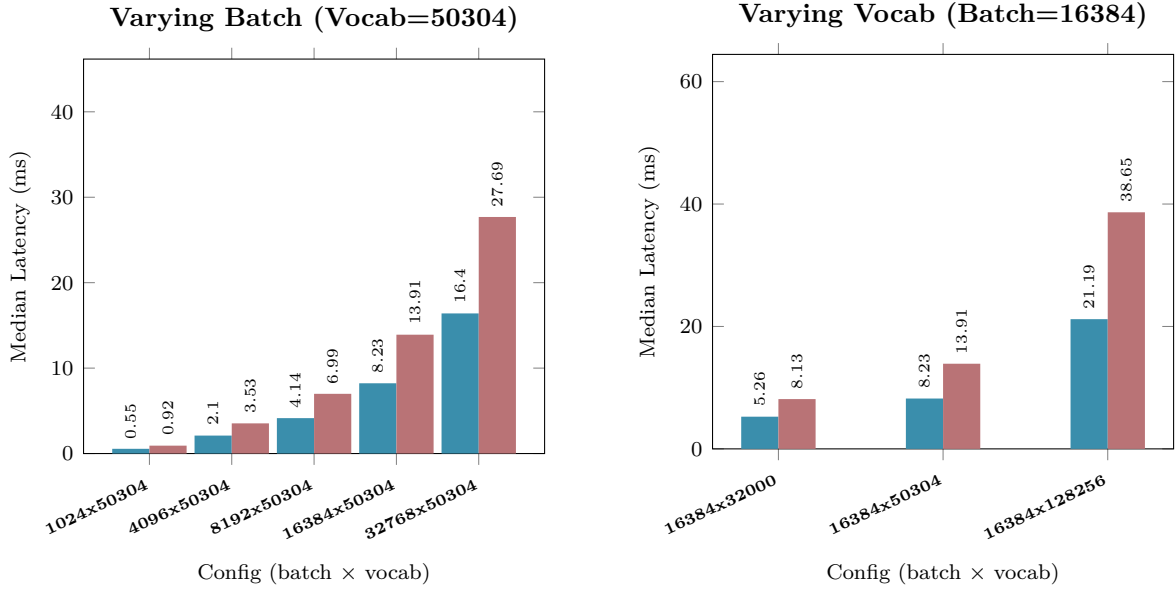

    \centering
    \begin{subfigure}{0.48\textwidth}
        \centering
        %
    \end{subfigure}
    \caption{Backward Loss median latency. Lower is better.}
\end{figure}

\vfill
\begin{figure}[H]
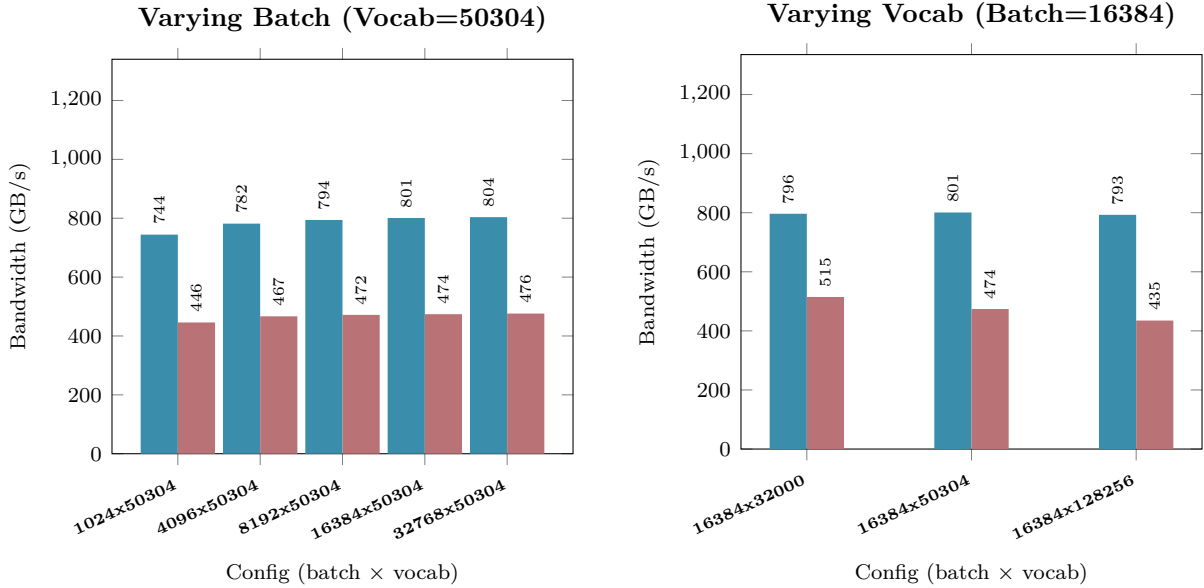

    \centering
    \begin{subfigure}{0.48\textwidth}
        \centering
        %
    \end{subfigure}
    \caption{Backward Loss memory bandwidth. Higher is better.}
\end{figure}

\vfill
\clearpage

\subsection{Reduce Sum Kernel \small (\textcolor{bluteal}{\rule{1.2ex}{1.2ex}} BluBridge, \textcolor{ptrose}{\rule{1.2ex}{1.2ex}} PyTorch)}
\label{app:reduce}

\begin{figure}[H]
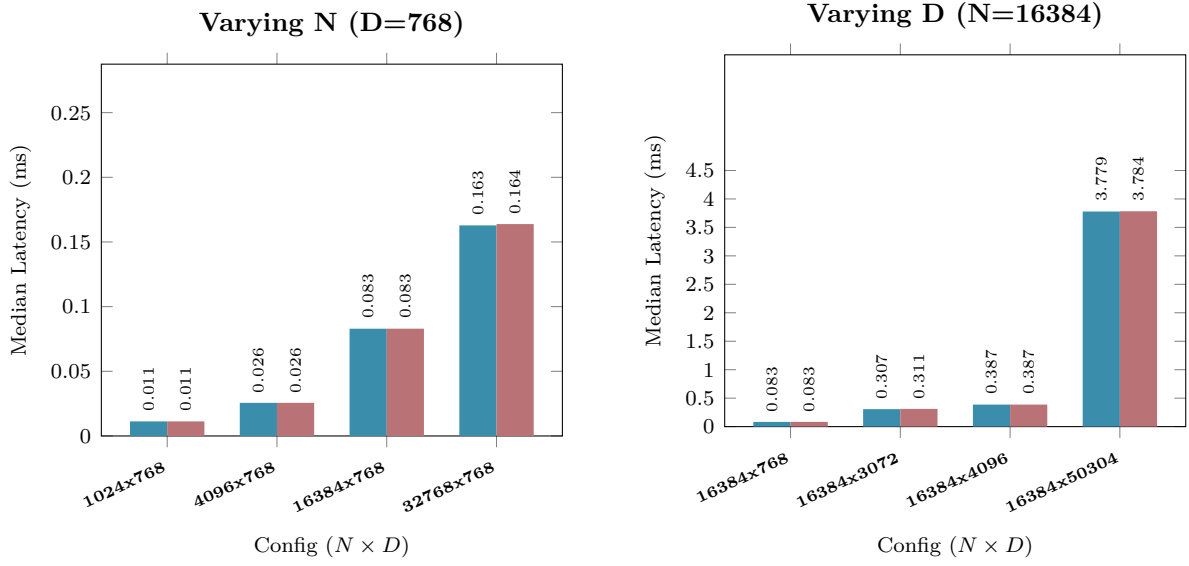

    \centering
    \begin{subfigure}{0.48\textwidth}
        \centering
        %
    \end{subfigure}
    \caption{Reduce Sum median latency. Lower is better.}
\end{figure}

\begin{figure}[H]
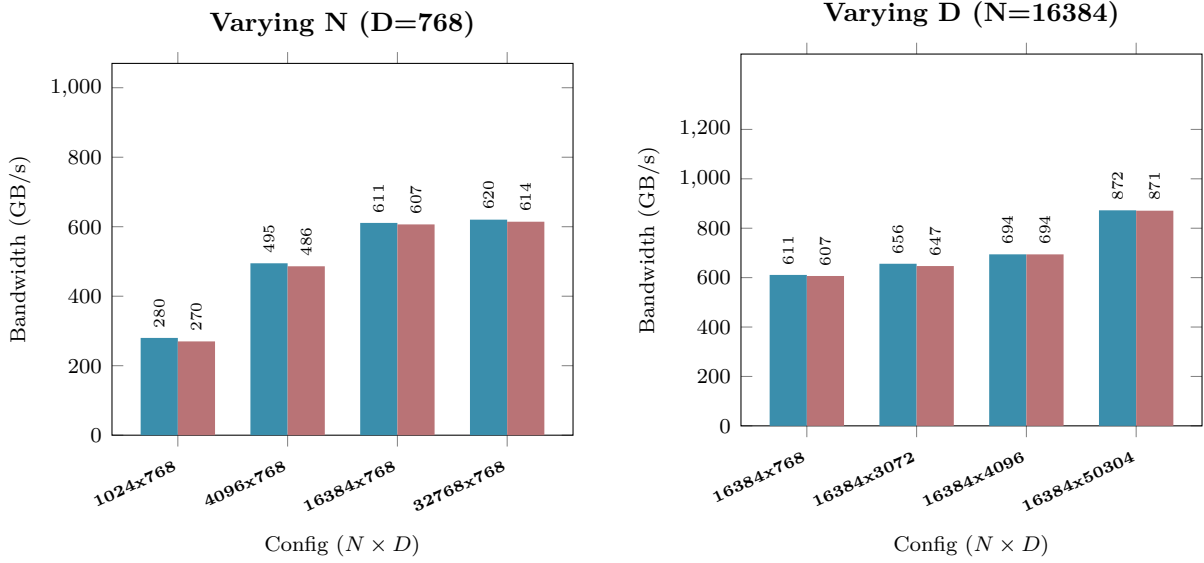

    \centering
    \begin{subfigure}{0.48\textwidth}
        \centering
        %
    \end{subfigure}
    \caption{Reduce Sum memory bandwidth. Higher is better.}
\end{figure}

\clearpage
\subsection{AdamW Optimizer Kernel \small (\textcolor{bluteal}{\rule{1.2ex}{1.2ex}} BluBridge, \textcolor{ptrose}{\rule{1.2ex}{1.2ex}} PyTorch)}
\label{app:adam}

\begin{figure}[H]
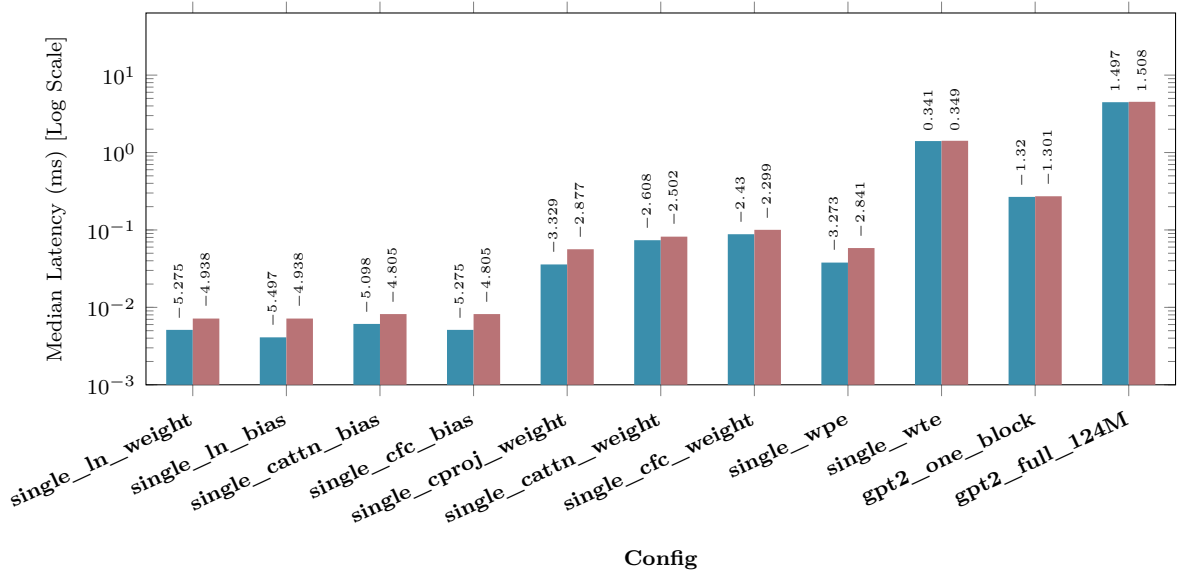

    \centering
    \begin{subfigure}{0.95\textwidth}
        \centering
        %
    \end{subfigure}
    \caption{AdamW Optimizer median latency. Lower is better.}
\end{figure}

\begin{figure}[H]
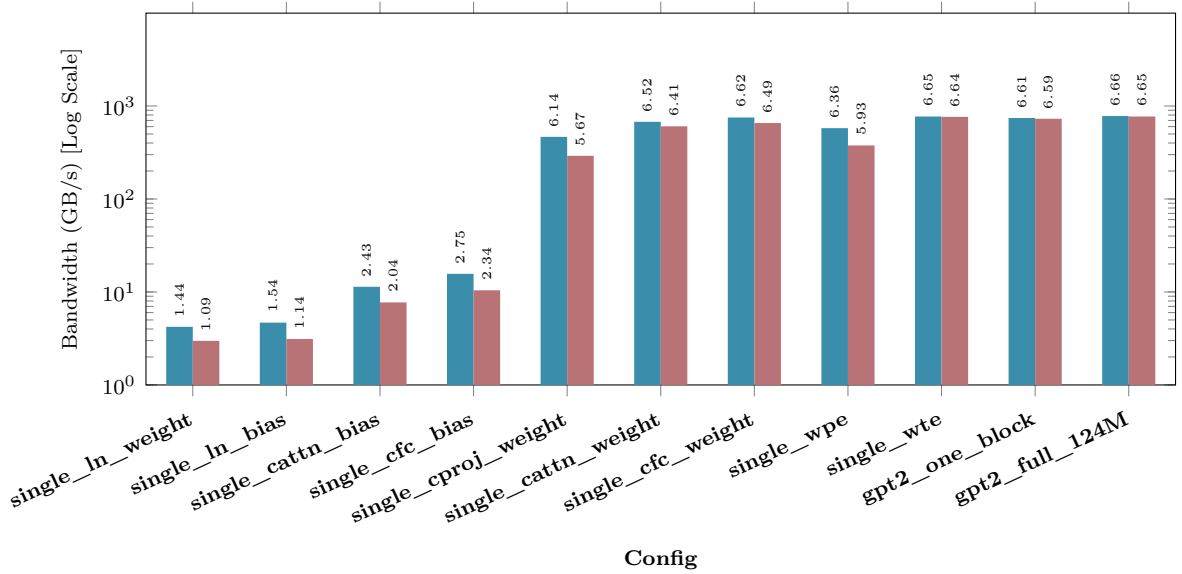

    \centering
    \begin{subfigure}{0.95\textwidth}
        \centering
        %
    \end{subfigure}
    \caption{AdamW Optimizer memory bandwidth. Higher is better.}
\end{figure}

\subsection{Distributed Data Parallel Execution}
\label{app:ddp}
The following distributed data-parallel benchmarks were evaluated on an \textbf{8x NVIDIA RTX 6000 Ada Lovelace} (48\,GB) multi-GPU hardware cluster, training a 124M-parameter GPT-2 configuration.
\begin{figure}[H]
    \centering
    \makebox[\textwidth]{\includegraphics[width=1.0\textwidth]{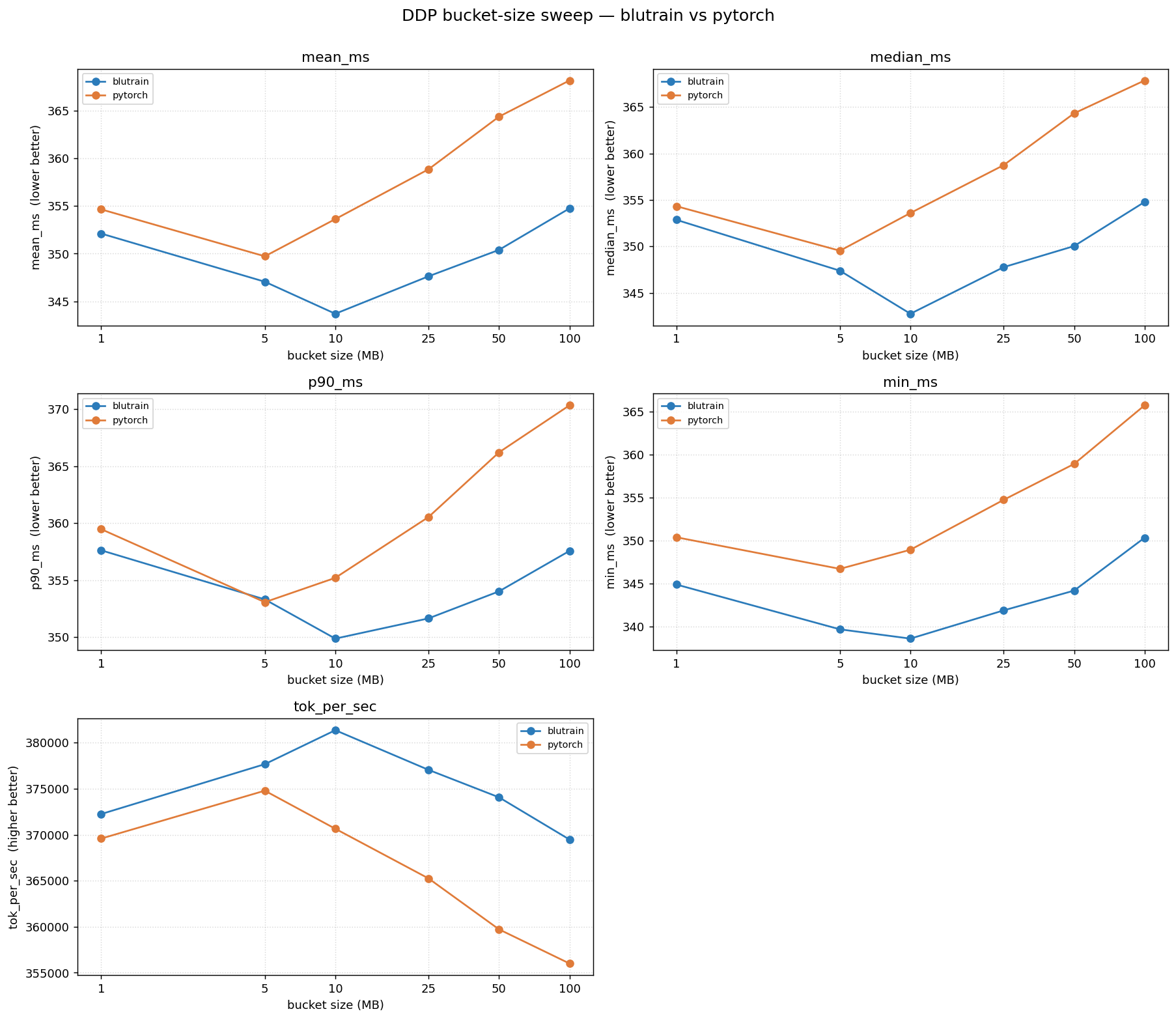}}
    \caption{Distributed Data Parallel (DDP) \code{AllReduce} latency and throughput scaling against varying bucket sizes (5MB to 100MB) during the training of a 124M-parameter GPT-2 configuration.}
\end{figure}

\subsection{Tensor Parallelism Benchmarks}
\label{app:tp}

The following benchmarks evaluate the exact execution characteristics of the tensor-parallel runtime. All empirical measurements were conducted during the training of a 124M-parameter GPT-2 configuration (FP32 precision, Global Batch: 524,288) on a dual \textbf{NVIDIA RTX 5070} GPU hardware topology.

\vspace*{\fill}
\begin{figure}[H]
    \centering
    \begin{subfigure}[t]{0.48\textwidth}
        \centering
        \begin{tikzpicture}
            \begin{axis}[
                ybar,
                bar width=35pt,
                width=5.9cm,
                height=5.6cm,
                scale only axis=true,
                enlarge x limits=0.8,
                enlarge y limits={upper, value=0.15},
                ylabel={Throughput (tok/s)},
                symbolic x coords={1 GPU, TP=2},
                xtick=\empty,
                nodes near coords={\pgfmathprintnumber[fixed,precision=0]{\pgfplotspointmeta}},
                every node near coord/.append style={font=\scriptsize\bfseries, color=black},
                ymin=0, ymax=45000,
                yticklabel style={/pgf/number format/fixed, /pgf/number format/precision=0, font=\scriptsize\bfseries},
                scaled y ticks=false,
                xticklabel style={font=\scriptsize\bfseries},
                ylabel style={font=\scriptsize},
                legend style={at={(0.5,1.0)}, anchor=south, draw=none, fill=none, font=\scriptsize},
                legend columns=-1, /tikz/every even column/.append style={column sep=0.25cm},
                legend image code/.code={\draw[#1, draw=none] (0cm,-0.1cm) rectangle (0.2cm,0.1cm);}
            ]
            \addplot[fill=ptrose, draw=none] coordinates {(1 GPU,25108)};
            \addplot[fill=bluteal, draw=none] coordinates {(TP=2,33961)};
            \legend{Baseline, BluTrain}
            \end{axis}
        \end{tikzpicture}
        \caption{Throughput Scaling: Distributing parameters and activations across two devices relieves memory pressure, allowing steady-state throughput to surge by 35.3\%.}
    \end{subfigure}
    \hfill
    \begin{subfigure}[t]{0.48\textwidth}
        \centering
        \begin{tikzpicture}
            \begin{axis}[
                ybar,
                bar width=35pt,
                width=5.9cm,
                height=5.6cm,
                scale only axis=true,
                enlarge x limits=0.8,
                enlarge y limits={upper, value=0.15},
                ylabel={Peak Memory (MiB)},
                symbolic x coords={1 GPU, TP=2},
                xtick=\empty,
                nodes near coords={\pgfmathprintnumber[fixed,precision=0]{\pgfplotspointmeta}},
                every node near coord/.append style={font=\scriptsize\bfseries, color=black},
                ymin=0, ymax=11000,
                yticklabel style={/pgf/number format/fixed, /pgf/number format/precision=0, font=\scriptsize\bfseries},
                scaled y ticks=false,
                ylabel style={font=\scriptsize},
                legend style={at={(0.5,1.0)}, anchor=south, draw=none, fill=none, font=\scriptsize},
                legend columns=-1, /tikz/every even column/.append style={column sep=0.25cm},
                legend image code/.code={\draw[#1, draw=none] (0cm,-0.1cm) rectangle (0.2cm,0.1cm);}
            ]
            \addplot[fill=ptrose, draw=none] coordinates {(1 GPU,8966)};
            \addplot[fill=bluteal, draw=none] coordinates {(TP=2,5762)};
            \legend{Baseline, BluTrain}
            \end{axis}
        \end{tikzpicture}
        \caption{Memory Pressure: Tensor Parallelism structurally halves the parameter footprints, dropping peak VRAM consumption from 8,966 MiB to 5,762 MiB per device.}
    \end{subfigure}
\end{figure}

\vfill
\begin{figure}[H]
    \centering
    \begin{subfigure}[t]{0.48\textwidth}
        \centering
        \begin{tikzpicture}
            \begin{axis}[
                ybar,
                bar width=35pt,
                width=5.9cm,
                height=5.6cm,
                scale only axis=true,
                enlarge x limits=0.8,
                enlarge y limits={upper, value=0.15},
                ylabel={Throughput (tok/s)},
                symbolic x coords={Megatron, BluTrain},
                xtick=\empty,
                nodes near coords={\pgfmathprintnumber[fixed,precision=0]{\pgfplotspointmeta}},
                every node near coord/.append style={font=\scriptsize\bfseries, color=black},
                ymin=0, ymax=45000,
                yticklabel style={/pgf/number format/fixed, /pgf/number format/precision=0, font=\scriptsize\bfseries},
                scaled y ticks=false,
                xticklabel style={font=\scriptsize\bfseries},
                ylabel style={font=\scriptsize},
                legend style={at={(0.5,1.0)}, anchor=south, draw=none, fill=none, font=\scriptsize},
                legend columns=-1, /tikz/every even column/.append style={column sep=0.25cm},
                legend image code/.code={\draw[#1, draw=none] (0cm,-0.1cm) rectangle (0.2cm,0.1cm);}
            ]
            \addplot[fill=ptrose, draw=none] coordinates {(Megatron,28374)};
            \addplot[fill=bluteal, draw=none] coordinates {(BluTrain,33961)};
            \legend{Megatron, BluTrain}
            \end{axis}
        \end{tikzpicture}
        \caption{Reference Comparison: At identical $TP=2$ degrees, BluTrain yields a 19.7\% advantage over Megatron-LM.}
    \end{subfigure}
    \hfill
    \begin{subfigure}[t]{0.48\textwidth}
        \centering
        \begin{tikzpicture}
            \begin{axis}[
                ybar,
                bar width=25pt,
                width=5.9cm,
                height=5.6cm,
                scale only axis=true,
                enlarge x limits=0.7,
                enlarge y limits={upper, value=0.25},
                ylabel={Throughput (tok/s)},
                symbolic x coords={Base TP, Async N=2, Async N=4},
                xtick=\empty,
                nodes near coords={\pgfmathprintnumber[fixed,precision=0]{\pgfplotspointmeta}},
                every node near coord/.append style={font=\scriptsize\bfseries, color=black},
                ymin=0, ymax=48000,
                yticklabel style={/pgf/number format/fixed, /pgf/number format/precision=0, font=\scriptsize\bfseries},
                scaled y ticks=false,
                ylabel style={font=\scriptsize},
                legend style={at={(0.5,1.0)}, anchor=south, draw=none, fill=none, font=\scriptsize},
                legend columns=-1, /tikz/every even column/.append style={column sep=0.25cm},
                legend image code/.code={\draw[#1, draw=none] (0cm,-0.1cm) rectangle (0.2cm,0.1cm);}
            ]
            \addplot[fill=ptrose, draw=none] coordinates {(Base TP,33961)};
            \addplot[fill=bluteal, draw=none] coordinates {(Async N=2,36562)};
            \addplot[fill=blugray, draw=none] coordinates {(Async N=4,36658)};
            \legend{Base TP, Async N=2, Async N=4}
            \end{axis}
        \end{tikzpicture}
        \caption{AsyncTP Throughput: all three bars are BluTrain at $TP{=}2$, where $N$ is the number of token chunks the step is split into; ``Base TP'' is BluTrain's own synchronous blocking-\code{AllReduce} baseline (our blocking run). Utilizing a dual-stream architecture, the asynchronous overlap protocol lifts steady-state performance by an additional $\sim$7.9\% over this baseline.}
    \end{subfigure}
\end{figure}

\vfill
\clearpage

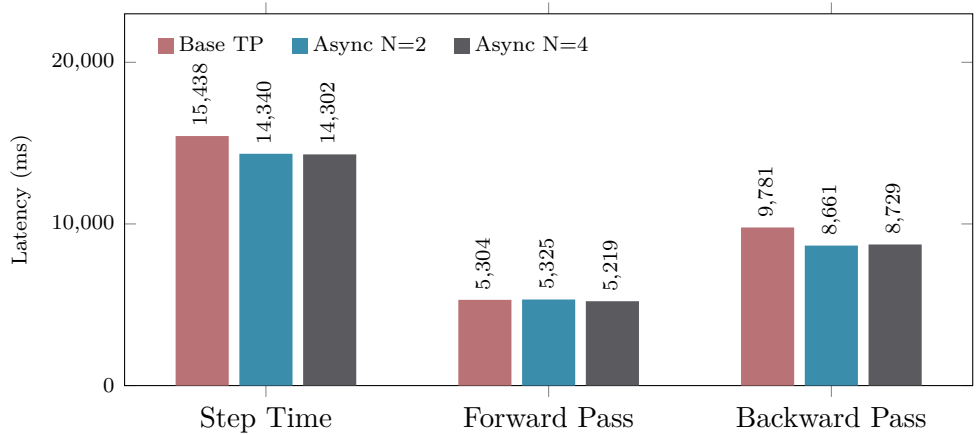
\begin{figure}[H]
    \centering
    \begin{tikzpicture}
        \begin{axis}[
            ybar=4pt,
            bar width=20pt,
            width=0.8\textwidth,
            height=6.5cm,
            enlarge x limits=0.25,
            enlarge y limits={upper, value=0.15},
            ylabel={Latency (ms)},
            symbolic x coords={Step Time, Forward Pass, Backward Pass},
            xtick=data,
            nodes near coords={\pgfmathprintnumber[fixed,precision=0]{\pgfplotspointmeta}},
            every node near coord/.append style={font=\scriptsize\bfseries, color=black, rotate=90, anchor=west},
            ymin=0, ymax=20000,
            yticklabel style={/pgf/number format/fixed, /pgf/number format/precision=0, font=\scriptsize\bfseries},
            scaled y ticks=false,
            ylabel style={font=\scriptsize},
            legend pos=north west,
            legend style={font=\scriptsize, draw=none, fill=none, legend columns=3, /tikz/every even column/.append style={column sep=0.3cm}},
            legend image code/.code={\draw[#1, draw=none] (0cm,-0.1cm) rectangle (0.2cm,0.1cm);}
        ]
        \addplot[fill=ptrose, draw=none] coordinates {(Step Time,15438) (Forward Pass,5304) (Backward Pass,9781)};
        \addplot[fill=bluteal, draw=none] coordinates {(Step Time,14340) (Forward Pass,5325) (Backward Pass,8661)};
        \addplot[fill=blugray, draw=none] coordinates {(Step Time,14302) (Forward Pass,5219) (Backward Pass,8729)};
        \legend{Base TP, Async N=2, Async N=4}
        \end{axis}
    \end{tikzpicture}
    \caption{Per-Stage Latency Breakdown: This visualizes exactly where the AsyncTP protocol recovers idle time. By hiding the synchronous \code{AllReduce} network transfers strictly behind the matmul compute kernels, it selectively collapses the backward pass latency without affecting the forward pass.}
\end{figure}
\subsection{Context Parallelism Benchmarks}
\label{app:cp}

The following benchmarks evaluate the context-parallel runtime across its three ring-rotator variants (\textbf{All-to-All}, \textbf{P2P}, and \textbf{AllGather}), each measured against a PyTorch attention-style baseline (\code{PT\_TF32{=}1}) in full FP32 precision at identical model configuration and global batch. Measurements span two dual-GPU testbeds: \textbf{2$\times$ RTX 5070} and \textbf{2$\times$ RTX 6000 Ada} (48\,GB). Reported throughput is the steady-state average; loss is the minimum validation loss reached over the full schedule. The 44M GPT-2 run uses context $1024$ and global batch $65{,}536$; the 163M run uses context $1024$ and global batch $524{,}288$.

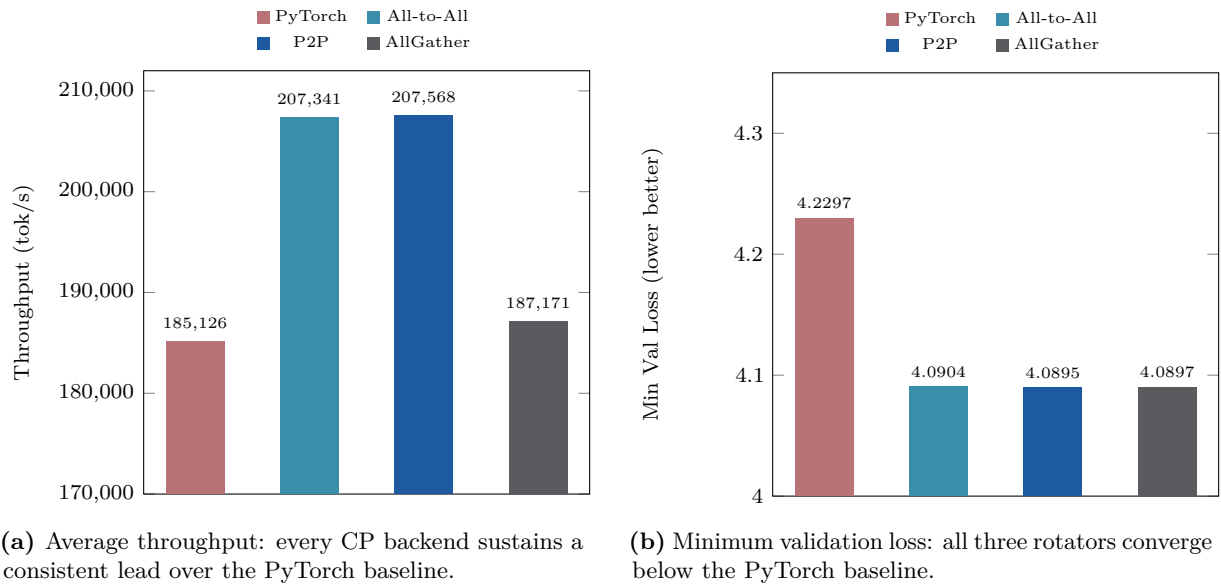
\begin{figure}[H]
    \centering
    \begin{subfigure}[t]{0.48\textwidth}
        \centering
        \begin{tikzpicture}
            \begin{axis}[
                ybar, bar width=22pt,
                width=5.9cm, height=5.6cm, scale only axis=true,
                enlarge x limits=0.15,
                ylabel={Throughput (tok/s)},
                symbolic x coords={PyTorch, All-to-All, P2P, AllGather},
                xtick=\empty,
                legend style={at={(0.5,1.02)}, anchor=south, legend columns=2, font=\tiny, draw=none, fill=none, /tikz/every even column/.append style={column sep=0.2cm}},
                legend image code/.code={\draw[#1,draw=none](0cm,-0.08cm)rectangle(0.16cm,0.08cm);},
                nodes near coords={\pgfmathprintnumber[fixed,precision=0]{\pgfplotspointmeta}},
                every node near coord/.append style={font=\tiny\bfseries, color=black},
                ymin=170000, ymax=212000,
                yticklabel style={/pgf/number format/fixed, font=\scriptsize, text width=1.05cm, align=right},
                scaled y ticks=false, ylabel style={font=\scriptsize},
            ]
            \addplot[fill=ptrose, draw=none, bar shift=0pt] coordinates {(PyTorch,185126)};
            \addplot[fill=bluteal, draw=none, bar shift=0pt] coordinates {(All-to-All,207341)};
            \addplot[fill=blublue, draw=none, bar shift=0pt] coordinates {(P2P,207568)};
            \addplot[fill=blugray, draw=none, bar shift=0pt] coordinates {(AllGather,187171)};
            \legend{PyTorch, All-to-All, P2P, AllGather}
            \end{axis}
        \end{tikzpicture}
        \caption{Average throughput: every CP backend sustains a consistent lead over the PyTorch baseline.}
    \end{subfigure}
    \hfill
    \begin{subfigure}[t]{0.48\textwidth}
        \centering
        \begin{tikzpicture}
            \begin{axis}[
                ybar, bar width=22pt,
                width=5.9cm, height=5.6cm, scale only axis=true,
                enlarge x limits=0.15,
                ylabel={Min Val Loss (lower better)},
                symbolic x coords={PyTorch, All-to-All, P2P, AllGather},
                xtick=\empty,
                legend style={at={(0.5,1.02)}, anchor=south, legend columns=2, font=\tiny, draw=none, fill=none, /tikz/every even column/.append style={column sep=0.2cm}},
                legend image code/.code={\draw[#1,draw=none](0cm,-0.08cm)rectangle(0.16cm,0.08cm);},
                nodes near coords={\pgfmathprintnumber[fixed,fixed zerofill,precision=4]{\pgfplotspointmeta}},
                every node near coord/.append style={font=\tiny\bfseries, color=black},
                ymin=4.0, ymax=4.35,
                yticklabel style={/pgf/number format/fixed, /pgf/number format/precision=2, font=\scriptsize, text width=1.05cm, align=right},
                scaled y ticks=false, ylabel style={font=\scriptsize},
            ]
            \addplot[fill=ptrose, draw=none, bar shift=0pt] coordinates {(PyTorch,4.2297)};
            \addplot[fill=bluteal, draw=none, bar shift=0pt] coordinates {(All-to-All,4.0904)};
            \addplot[fill=blublue, draw=none, bar shift=0pt] coordinates {(P2P,4.0895)};
            \addplot[fill=blugray, draw=none, bar shift=0pt] coordinates {(AllGather,4.0897)};
            \legend{PyTorch, All-to-All, P2P, AllGather}
            \end{axis}
        \end{tikzpicture}
        \caption{Minimum validation loss: all three rotators converge below the PyTorch baseline.}
    \end{subfigure}
    \caption{Context Parallelism on \textbf{44M GPT-2, dual RTX 5070}: the CP backends preserve a throughput lead over PyTorch alongside an equal-or-better convergence.}
\end{figure}

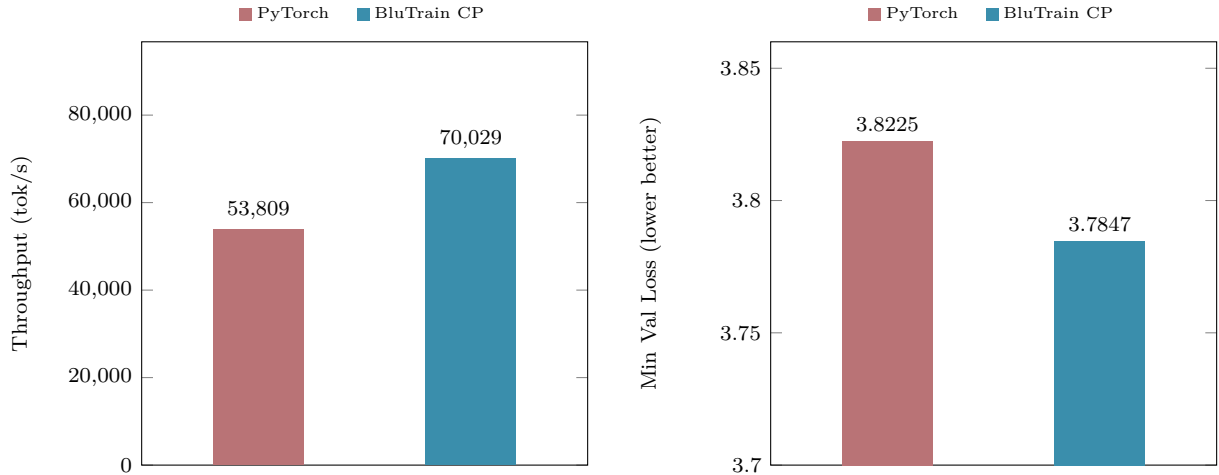
\begin{figure}[H]
    \centering
    \begin{subfigure}[t]{0.48\textwidth}
        \centering
        \begin{tikzpicture}
            \begin{axis}[
                ybar, bar width=34pt,
                width=5.9cm, height=5.6cm, scale only axis=true,
                enlarge x limits=0.55,
                enlarge y limits={upper, value=0.18},
                ylabel={Throughput (tok/s)},
                symbolic x coords={PyTorch, BluTrain CP},
                xtick=\empty,
                legend style={at={(0.5,1.02)}, anchor=south, legend columns=-1, font=\tiny, draw=none, fill=none, /tikz/every even column/.append style={column sep=0.3cm}},
                legend image code/.code={\draw[#1,draw=none](0cm,-0.08cm)rectangle(0.16cm,0.08cm);},
                nodes near coords={\pgfmathprintnumber[fixed,precision=0]{\pgfplotspointmeta}},
                every node near coord/.append style={font=\scriptsize\bfseries, color=black},
                ymin=0, ymax=82000,
                yticklabel style={/pgf/number format/fixed, font=\scriptsize, text width=1.05cm, align=right},
                scaled y ticks=false, ylabel style={font=\scriptsize},
            ]
            \addplot[fill=ptrose, draw=none, bar shift=0pt] coordinates {(PyTorch,53809)};
            \addplot[fill=bluteal, draw=none, bar shift=0pt] coordinates {(BluTrain CP,70029)};
            \legend{PyTorch, BluTrain CP}
            \end{axis}
        \end{tikzpicture}
        \caption{Average throughput: BluTrain CP delivers a \textbf{+30.1\%} gain over the PyTorch baseline.}
    \end{subfigure}
    \hfill
    \begin{subfigure}[t]{0.48\textwidth}
        \centering
        \begin{tikzpicture}
            \begin{axis}[
                ybar, bar width=34pt,
                width=5.9cm, height=5.6cm, scale only axis=true,
                enlarge x limits=0.55,
                ylabel={Min Val Loss (lower better)},
                symbolic x coords={PyTorch, BluTrain CP},
                xtick=\empty,
                legend style={at={(0.5,1.02)}, anchor=south, legend columns=-1, font=\tiny, draw=none, fill=none, /tikz/every even column/.append style={column sep=0.3cm}},
                legend image code/.code={\draw[#1,draw=none](0cm,-0.08cm)rectangle(0.16cm,0.08cm);},
                nodes near coords={\pgfmathprintnumber[fixed,fixed zerofill,precision=4]{\pgfplotspointmeta}},
                every node near coord/.append style={font=\scriptsize\bfseries, color=black},
                ymin=3.7, ymax=3.86,
                yticklabel style={/pgf/number format/fixed, /pgf/number format/precision=2, font=\scriptsize, text width=1.05cm, align=right},
                scaled y ticks=false, ylabel style={font=\scriptsize},
            ]
            \addplot[fill=ptrose, draw=none, bar shift=0pt] coordinates {(PyTorch,3.8225)};
            \addplot[fill=bluteal, draw=none, bar shift=0pt] coordinates {(BluTrain CP,3.7847)};
            \legend{PyTorch, BluTrain CP}
            \end{axis}
        \end{tikzpicture}
        \caption{Minimum validation loss: CP converges below the PyTorch baseline at the same step budget.}
    \end{subfigure}
    \caption{Context Parallelism on \textbf{163M GPT-2}, dual RTX 6000 Ada: higher end-to-end throughput than PyTorch at an equal-or-better validation loss.}
\end{figure}

Across both testbeds the context-parallel runtime delivers higher end-to-end throughput than the PyTorch baseline while matching or improving the final validation loss, with the largest margin on the RTX 6000 Ada testbed.

\begin{table}[H]
\centering
\caption{Peak GPU memory footprint under context-parallel training with BluTrain's caching-allocator optimization, BluTrain vs PyTorch.}
\label{tab:cp_mem}
\small
\setlength{\tabcolsep}{8pt}
\begin{tabular}{lrrr}
\toprule
Model & BluTrain (GB) & PyTorch (GB) & Reduction \\
\midrule
GPT-2 44M & \textbf{2.606} & 2.872 & 9.3\% \\
GPT-2 25M & \textbf{2.514} & 2.798 & 10.2\% \\
\bottomrule
\end{tabular}
\end{table}

\subsection{Checkpointing Benchmarks}
\label{app:ckpt}

These benchmarks evaluate the checkpointing subsystem of the orchestration layer, which serialises the full training state (parameters, optimizer moments, RNG state, step, epoch, and loss) into a single self-describing file. BluTrain exposes a synchronous path and an asynchronous path that snapshots the state into pinned host memory and writes it on a background thread; PyTorch's \code{torch.save}/\code{torch.load} offers no native asynchronous path. All figures are median latencies in full FP32 with end-to-end integrity verification, measured on the 44M (532\,MB checkpoint) and 124M (1.49\,GB checkpoint) GPT-2 configurations. For the asynchronous paths, the reported value is the staging stall, i.e.\ the only portion the training loop blocks on.

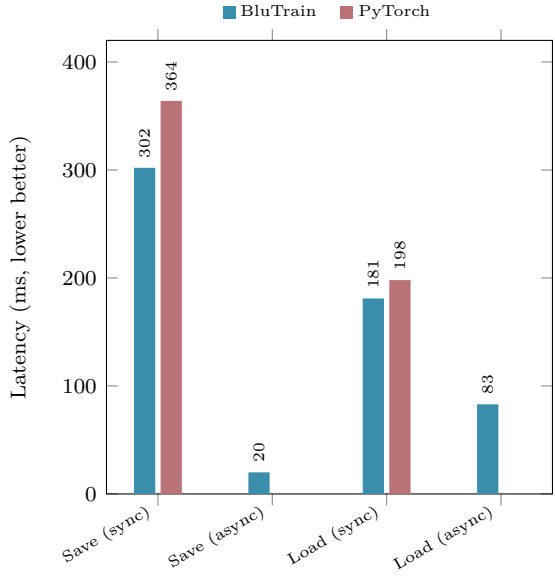
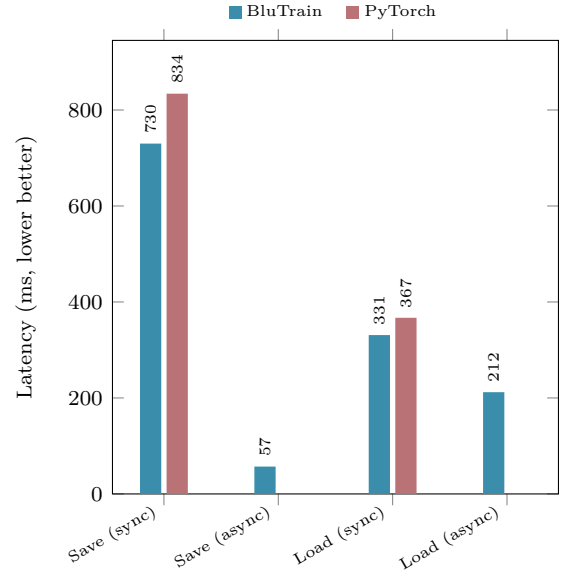
\begin{figure}[H]
    \centering
    \begin{subfigure}[t]{0.48\textwidth}
        \centering
        \begin{tikzpicture}
            \begin{axis}[
                ybar, bar width=8pt,
                width=5.9cm, height=6cm, scale only axis=true,
                enlarge x limits=0.15,
                ylabel={Latency (ms, lower better)},
                symbolic x coords={Save (sync), Save (async), Load (sync), Load (async)},
                xtick=data,
                xticklabel style={font=\tiny, rotate=30, anchor=east},
                nodes near coords={\pgfmathprintnumber[fixed,precision=0]{\pgfplotspointmeta}},
                every node near coord/.append style={font=\tiny, color=black, rotate=90, anchor=west},
                yticklabel style={/pgf/number format/fixed, font=\scriptsize, text width=0.6cm, align=right},
                scaled y ticks=false, ylabel style={font=\scriptsize},
                legend style={at={(0.5,1.02)}, anchor=south, legend columns=-1, font=\tiny, draw=none, fill=none, /tikz/every even column/.append style={column sep=0.25cm}},
                legend image code/.code={\draw[#1,draw=none](0cm,-0.08cm)rectangle(0.16cm,0.08cm);},
                ymin=0, ymax=420,
            ]
            \addplot[fill=bluteal, draw=none] coordinates {(Save (sync),302) (Save (async),20) (Load (sync),181) (Load (async),83)};
            \addplot[fill=ptrose, draw=none] coordinates {(Save (sync),364) (Load (sync),198)};
            \legend{BluTrain, PyTorch}
            \end{axis}
        \end{tikzpicture}
        \caption{GPT-2 44M (532\,MB checkpoint).}
    \end{subfigure}
    \hfill
    \begin{subfigure}[t]{0.48\textwidth}
        \centering
        \begin{tikzpicture}
            \begin{axis}[
                ybar, bar width=8pt,
                width=5.9cm, height=6cm, scale only axis=true,
                enlarge x limits=0.15,
                ylabel={Latency (ms, lower better)},
                symbolic x coords={Save (sync), Save (async), Load (sync), Load (async)},
                xtick=data,
                xticklabel style={font=\tiny, rotate=30, anchor=east},
                nodes near coords={\pgfmathprintnumber[fixed,precision=0]{\pgfplotspointmeta}},
                every node near coord/.append style={font=\tiny, color=black, rotate=90, anchor=west},
                yticklabel style={/pgf/number format/fixed, font=\scriptsize, text width=0.6cm, align=right},
                scaled y ticks=false, ylabel style={font=\scriptsize},
                legend style={at={(0.5,1.02)}, anchor=south, legend columns=-1, font=\tiny, draw=none, fill=none, /tikz/every even column/.append style={column sep=0.25cm}},
                legend image code/.code={\draw[#1,draw=none](0cm,-0.08cm)rectangle(0.16cm,0.08cm);},
                ymin=0, ymax=945,
            ]
            \addplot[fill=bluteal, draw=none] coordinates {(Save (sync),730) (Save (async),57) (Load (sync),331) (Load (async),212)};
            \addplot[fill=ptrose, draw=none] coordinates {(Save (sync),834) (Load (sync),367)};
            \legend{BluTrain, PyTorch}
            \end{axis}
        \end{tikzpicture}
        \caption{GPT-2 124M (1.49\,GB checkpoint).}
    \end{subfigure}
    \caption{Single-file checkpoint latency, BluTrain vs PyTorch (median, FP32, integrity-verified). Synchronous save and load are directly comparable across systems; the asynchronous bars report only the staging stall the training loop actually waits on, for which PyTorch has no equivalent. The full asynchronous save completes in 177\,ms for 44M (20\,ms GPU stall + 157\,ms background disk write, hidden) and 503\,ms for 124M (57\,ms GPU stall + 446\,ms disk write, hidden).}
\end{figure}

\end{document}